\documentclass[letterpaper]{article} 
\usepackage{aaai23}  
\usepackage{times}  
\usepackage{xcolor}
\usepackage{helvet}  
\usepackage{courier}  
\usepackage[hyphens]{url}  
\usepackage{graphicx} 
\def\UrlFont{\rm}  
\usepackage{natbib}  
\usepackage{caption} 
\usepackage{subcaption}
\usepackage{algorithm}
\usepackage{algorithmic}

\usepackage{newfloat}
\usepackage{listings}
\DeclareCaptionStyle{ruled}{labelfont=normalfont,labelsep=colon,strut=off} 
\floatstyle{ruled}
\newfloat{listing}{tb}{lst}{}
\floatname{listing}{Listing}
\title{Inapplicable Actions Learning for Knowledge Transfer in Reinforcement Learning}
\author {
    Leo Ardon,
    Alberto Pozanco,
    Daniel Borrajo,
    Sumitra Ganesh
}
\affiliations{
J.P. Morgan AI Research \\
\{leo.ardon,alberto.pozancolancho, daniel.borrajo, sumitra.ganesh\}@jpmorgan.com
}

\usepackage{amsmath,amsfonts,bm}

\def\figref#1{figure~\ref{#1}}

\def\secref#1{section~\ref{#1}}

\def\eqref#1{equation~\ref{#1}}

\def\1{\bm{1}}

\def\vm{{\bm{m}}}

\def\vs{{\bm{s}}}

\def\evm{{m}}

\DeclareMathAlphabet{\mathsfit}{\encodingdefault}{\sfdefault}{m}{sl}
\SetMathAlphabet{\mathsfit}{bold}{\encodingdefault}{\sfdefault}{bx}{n}

\def\sA{{\mathbb{A}}}

\def\sD{{\mathbb{D}}}

\def\sI{{\mathbb{I}}}

\def\sR{{\mathbb{R}}}
\def\sS{{\mathbb{S}}}

\newcommand{\E}{\mathbb{E}}
\newcommand{\Ls}{\mathcal{L}}

\newtheorem{definition}{Definition}[section]

\newcommand{\defref}[1]{Definition~\ref{#1}}
\newcommand{\algoref}[1]{Algorithm~\ref{#1}}
\newcommand{\algolineref}[1]{\texttt{l.\ref{#1}}}
\newcommand{\appendixref}[1]{Appendix~\ref{#1}}
\renewcommand{\secref}[1]{Section~\ref{#1}}
\renewcommand{\figref}[1]{Figure~\ref{#1}}

\def\bheader{\begin{tabular}{l}}
\def\eheader{\end{tabular}}

\begin{document}

\maketitle

\begin{abstract}
Reinforcement Learning (RL) algorithms are known to scale poorly to environments with many available actions, requiring numerous samples to learn an optimal policy. The traditional approach of considering the same fixed action space in every possible state implies that the agent must understand, while also learning to maximize its reward, to ignore irrelevant actions such as $\textit{inapplicable actions}$ (i.e.~actions that have no effect on the environment when performed in a given state). 
Knowing this information can help reduce the sample complexity of RL algorithms by masking the inapplicable actions from the policy distribution to only explore actions relevant to finding an optimal policy. While this technique has been formalized for quite some time within the Automated Planning community with the concept of precondition in the STRIPS language, RL algorithms have never formally taken advantage of this information to prune the search space to explore. This is typically done in an ad-hoc manner with hand-crafted domain logic added to the RL algorithm. 
In this paper, we propose a more systematic approach to introduce this knowledge into the algorithm. We (i) standardize the way knowledge can be manually specified to the agent; and (ii) present a new framework to autonomously learn the partial action model encapsulating the precondition of an action jointly with the policy. 
We show experimentally that learning inapplicable actions greatly improves the sample efficiency of the algorithm by providing a reliable signal to mask out irrelevant actions. Moreover, we demonstrate that thanks to the transferability of the knowledge acquired, it can be reused in other tasks and domains to make the learning process more efficient.
\end{abstract}

\section{Introduction}


The field of Deep Reinforcement Learning (DRL), using neural networks as function approximators in Reinforcement Learning (RL) algorithms, has seen many successes in recent years~\citep{Mnih_2015,silver2016mastering,Vinyals_2017}. Despite the impressive results this learning technique has shown, DRL is often criticized for being data hungry and sample inefficient.

Incorporating domain knowledge has been shown to be a valid approach to increase the learning effectiveness of DRL algorithms~\citep{Vinyals_2017,kool2018attention,Even-Dar_Mannor_Mansour_2006,pmlr-v80-barreto18a, Fernandez_Garcia_Veloso_2010,spooner2021factored}. We focus our attention to a specific type of knowledge in the form of \textit{inapplicable actions}, i.e. actions that do not modify the environment when performed in a particular state.
While this knowledge is explicitly defined in other types of planning systems, as preconditions of actions~\cite{ghallab2004automated}, it is rarely used within the RL community.
This information that can be interpreted as the \textit{rules of the game}, can help reduce the sample complexity of RL algorithms by pruning irrelevant state-action pairs from the search space the RL agent needs to explore.
This \textit{action masking} technique contributed to several breakthroughs in the field~\citep{Vinyals_2017,kool2018attention}. Unfortunately, a systematic way to incorporate this type of knowledge into the RL algorithm seems to be lacking from the literature.

For cases where the applicability of an action in a given state is unknown, we propose to turn one of RL algorithms’ weaknesses into an advantage and use the data collected by the RL agent through exploration, to train a partial action model able to identify whether an action is (in)applicable in a given state. 
As the model learns to identify the features encoding the constraints of the environment, it provides an increasingly accurate signal to the RL algorithm to mask out inapplicable actions, leading to a more sample efficient algorithm.

By learning the task-agnostic constraints, this new component not only helps improving the agent's learning process, but it also encapsulates knowledge about the environment in a manner that is both interpretable and transferable.
The information acquired solving a particular task can be shared to new tasks in the same domain, as well as to other similar domains that share common features. Instead of solving the new problem from scratch, the RL algorithm makes use of the trained classifier to limit the exploration of inapplicable actions.

In this work, we direct our efforts to improve the sample efficiency of Policy Gradient algorithms that have been proven to converge when action masking is used~\citep{huang2022closer}.
Our main contributions are: \textbf{(i)} incorporate an explicit representation of actions constraints in the Options Framework~\citep{SUTTON1999181} with the definition of an applicability function; \textbf{(ii)} show how various levels of hand-coded knowledge about the environment can significantly decrease the time required to learn a policy; \textbf{(iii)} provide an algorithm to jointly train a policy and an inapplicable actions classifier and empirically show that our method is more sample efficient than standard Policy Gradient algorithms; and \textbf{(iv)} present a new transfer learning technique for RL, where the knowledge previously acquired about the constraints of the environment is shared across tasks and domains by reusing a trained inapplicable actions classifier.

\section{Background}

We consider the $T$-episodic Reinforcement Learning problem in a Markov Decision Process with a discrete actions space, denoted as $\langle \sS, \sA, P(s, a, s'), R(s, a, s'), T, \mu_0, \gamma \rangle$. $\sS$ is the state space, $\sA$ the discrete action space, $P: \sS \times \sA \times \sS \rightarrow [0, 1]$ the Markovian state transition probability, $R: \sS \times \sA \times \sS \rightarrow [0, 1]$ the reward function, $T$ the maximum episode length, $\mu_0$ the initial state distribution, and $\gamma$ the discount factor. A stochastic policy $\pi_{\theta}: \sS \times \sA \rightarrow [0, 1]$ characterizes a function, parametrized by $\theta$, assigning a probability to an action given a state. The objective of an RL agent is to learn the optimal policy $\pi_{\theta}$ such that its expected discounted return is maximized.
\begin{align*}
    J &= \E_{\substack{s_0 \sim \mu_0 \\ a_t \sim \pi_{\theta}(\cdot | s_t) \\ s_{t+1} \sim P(\cdot | s_t, a_t)}} \bigg[ \sum^{T-1}_{t=0} \gamma^t R(s_t, a_t, s_{t+1}) \bigg] \\
    &= \E_{\mu_0, \pi_{\theta}} \bigg[ \sum^{T-1}_{t=0} \gamma^t R(s_t, a_t, s_{t+1}) \bigg]
\end{align*}

Where $s_{t+1}$ is sampled from the state transition probability distribution $P(\cdot|s_t, a_t)$, $a_t$ from the policy $\pi_{\theta}(\cdot|s_t)$ and $s_0$ from the initial state distribution $\mu_0$.


%

The family of Policy Gradient algorithms, initially introduced in~\citet{Sutton_McAllester_Singh_Mansour_1999}, aims at finding the optimal value for parameter $\theta$ such that the resulting policy generates the maximum expected returns. The parameter $\theta$ is updated in a gradient ascent fashion $\theta_{i+1} = \theta_{i} + \alpha \nabla_{\theta} J_i$ where the policy gradient $\nabla_{\theta} J$ has been shown to take the form:
\begin{gather} \label{eq:policy_gradient}
\nabla_{\theta} J = \E_{\mu_0, \pi_{\theta}} \big[ A^{\pi_{\theta}}_t \nabla_{\theta} \log \pi_{\theta}(a_t | s_t) \big]; \\
\text{ with } A^{\pi_{\theta}}_t = Q^{\pi_{\theta}}(s_t, a_t) - V^{\pi_{\theta}}(s_t) \notag \\
Q^{\pi_{\theta}}(s, a) = \E_{\pi_{\theta}} \bigg[ \sum^{T-1}_{k=0} \gamma^{k} R(s_{t+k}, a_{t+k}, s_{t+k+1}) \big| s_t=s, a_t=a \bigg] \notag\\
V^{\pi_{\theta}}(s) = \E_{\pi_{\theta}} \bigg[ \sum^{T-1}_{k=0} \gamma^{k} R(s_{t+k}, a_{t+k}, s_{t+k+1}) \big| s_t=s \bigg] \notag
\end{gather}
$V^{\pi_{\theta}}(s)$ is the value function and represents the expected returns received starting in state $s$ and following the policy $\pi_{\theta}$. Similarly, $Q^{\pi_{\theta}}(s, a)$ is the action value function and indicates the expected returns when action $a$ is taken in state $s$ and the policy $\pi_{\theta}$ is followed thereafter.




\section{Masking Inapplicable Actions}

In many real-world problems, the set of actions that an agent can perform varies depending on the state of the environment: $\sA(\vs)$. These variations can be explained by multiple factors such as expiration of resources (an agent running out of inventory will no longer be able to sell its product), the rules of the game (not being able to move its rook diagonally when playing chess), etc. Despite being supported by the standard Markov Decision Process framework, the ability for each state $\vs$ to have its own set of feasible actions seems to have been mainly ignored by standard RL algorithms. Their implementations typically assume a fixed set, concatenating all the possible actions that an agent can execute throughout the episode. 
We focus our attention to \textit{inapplicable actions} (\defref{def:inapplicable_action}), which are actions that have no effect on the environment when executed in a given state~\footnote{These are different from forbidden or terminal actions typically implemented with reward shaping or by interrupting the episode abruptly.}.

\begin{definition} \label{def:inapplicable_action} Inapplicable action

Given a T-episodic MDP defined as $\langle \sS, \sA, P(\vs, a, \vs'), R(\vs, a, \vs'), T, \mu_0, \gamma \rangle$ and a distance measure between two states: $d(s, s')$, an action is said to be \textbf{\emph{inapplicable in a given state}} if the distance between the state of the environment before and after the action was taken is lower than a small value $\varepsilon$. We use $\sI(\vs)$ to denote the set of inapplicable actions in state $\vs$.

\begin{equation} \label{eq:inapplicable_action}
    \forall a \in \sI(\vs) \subseteq \sA ; \text{P}(d(\vs_{t+1},\vs_t) \leq \varepsilon |\vs_t=\vs, a_t=a) = 1
\end{equation}

\end{definition}

This formulation of \textit{inapplicable action} assumes a certain level of observability of the environment. While full observability is not necessary, the agent requires however to have visibility on the features of the environment that the action $a$ has effect on. Addressing this limitation could be the object of future work.

We can leverage the notion of \emph{inapplicable action} in the \emph{Option Framework} \citep{SUTTON1999181} and define the set of available options as one-step option $\omega_a = \langle \mathcal{I}_{\omega_a}, \pi_{\omega_a},  \beta_{\omega_a} \rangle$ for each of the primitive action $a \in \sA$ where:
\begin{itemize}
    \item The initiation set $\mathcal{I}_{\omega_a} = \{ \vs \; : a \notin \sI(\vs) \; \forall \vs \in \sS \}$
    \item The option policy $\pi_{\omega_a}(\vs, a) = 1 \; \forall \vs \in \sS$
    \item The termination condition $\beta_{\omega_a}(\vs) = 1 \; \forall s \in \sS$
\end{itemize}

As the construction of the initiation set $\mathcal{I}$ can become costly in large environments, we introduce a new component $C(\vs, a)$, called the \textbf{applicability function}, returning on demand the probability of an action $a$ to be applicable in state $\vs$. The applicability function can be used to identify \textit{inapplicable actions} and determine whether an option $\omega_a$ can be ``initiated'' in a given state. This new component can help prune the search space of the problem and therefore improve the sample efficiency of RL algorithms.


We can use the \emph{applicability function} to define the initiation set of a given option $\omega_a$: 

\begin{gather}
\mathcal{I}_{\omega_a} = \{ \vs : C(s, a) \geq \tau \; \forall s \in \sS \} \\
\text{with } C(\vs, a) = P(a \notin \sI(\vs)) \notag
\end{gather}





The \emph{applicability function} also makes the definition of an option/action masking function $\vm: \sS \rightarrow \sR^{|\sA|}$ straightforward in the discrete action space case by simply calling the applicability function $C(\cdot, \cdot)$ for every element of the action space. As $C$ returns the probability of being applicable, we evaluate whether this value is below a certain cut-off threshold $\tau$ passed as parameter of the algorithm. This threshold is typically set to $0.5$ but can be adjusted depending on the problem, to make the identification of inapplicable actions more specific or more sensitive. The mask generated is thus a vector of $0$ for inapplicable actions, and $1$ for applicable actions that will be multiplied by the probability distribution returned by the policy over options $\pi_\theta$. The distribution is then re-normalized using the $\texttt{softmax}$ function\footnote{For most of the existing policy gradient algorithms, the output of the policy is the logit associated with each of the actions. In such case, the procedure consists of replacing the logit of the inapplicable actions by a very large negative number.}.
\begin{gather} \label{eq:masking_process}
    \pi^{\text{mask}}_{\theta}(\cdot | \vs, \pi_{\theta}, \vm(\vs)) = \texttt{softmax}(\vm(\vs) \cdot \pi_{\theta}(\cdot | \vs)) \\
    \text{with } \;\;
    m_i(\vs) = \begin{cases}
    0 & \text{ if }C(\vs, a_i) < \tau \notag \\
    1 & \text{otherwise}
    \end{cases} \;\; \forall i \in \{1..|\sA|\}
\end{gather}

In the following sections, we look at how the applicability function can be formalized to be used by Policy Gradient algorithms and how knowledge can be passed into the algorithm to help make the learning process more efficient.

\subsection{Inapplicable Actions Masking via Domain Knowledge}

The knowledge about inapplicable actions is, in some cases, directly associated with the rules specified by the environment (e.g the rules of the game) or simply common sense (e.g trying to drop an object that we do not hold). 
While it seems trivial for a human to use implicit knowledge previously acquired to explore only the relevant actions, RL algorithms do not have this information available by default.

People have started using the concept of actions masking to improve the learning efficiency of RL algorithm, but the rules are typically hand-coded in the algorithm itself \citep{Vinyals_2017,kool2018attention}, introducing environment specific logic into the RL algorithm.
Implementations such as RLLib Parametric Action Spaces \citep{rllib} takes one step towards abstracting away the action masking logic from the RL policy and propose to include the mask in the observation space. However, no consensus has been found on the best way to encode this information in a generic way.

Similarly to the concept of \emph{precondition} defined in the STRIPS language ~\citep{Fikes_Nilsson_1971} and used by the Automated Planning community, we argue that the \emph{applicability function} $C$ is domain specific and should therefore be part of the environment definition. 
We thus propose to extend the OpenAI Gym environment interface~\citep{Brockman_Cheung_Pettersson_Schneider_Schulman_Tang_Zaremba_2016} to include a new standardized method $\texttt{is\_applicable}$ returning whether an action is applicable in the current state of the environment. This method formalizing the applicability function $C$ of the SDAS-MDP will be called by the RL algorithm to create the actions mask to only sample applicable actions from the policy. 



\begin{algorithm}[h!]
\footnotesize
\caption{PseudoCode: Policy Gradient with inapplicable actions learning}
\label{alg:algorithm}
\textbf{Input}: policy parameter: $\theta_0$, classifier parameter: $\phi_0$, classifier exploration threshold: $\epsilon_0$, training flag: \texttt{train}\\
\textbf{Output}: trained policy parameters $\theta_K$, trained classifier parameters $\phi_K$
\begin{algorithmic}[1]
\STATE $\sD \leftarrow \{\}$ \COMMENT{instantiate the rollout buffer}
\FOR{$k = 1, 2, \ldots, K$}
\FOR{worker $= 1, 2, \ldots, N$}
\STATE \textsc{CollectTrajectories}($\pi_{\theta_k}$, $\mathcal{C}_{\phi_k}$, $\epsilon_k$, $\sD$). \COMMENT{with $\mathcal{C}_{\phi_k}$ the actions classifier}
\STATE Compute returns $\hat{R}_t$ and advantage estimates $\hat{A}_t$ using current value function $V_{k}$
\ENDFOR
\FOR{epochs=$1, 2, \ldots, M$} 
    \STATE Sample data from the replay buffer $\sD$
    \STATE Optimize $\Ls^{\text{Policy}}$ w.r.t $\theta$ via SGD with Adam optimizer
    \IF {\texttt{train}=\texttt{TRUE}}
        \STATE Balance applicable and inapplicable samples \COMMENT{balance dataset} \label{alg_line:balance_data}
        \STATE Optimize $\Ls^{\mathcal{C}}$ w.r.t $\phi$ via SGD with Adam optimizer \COMMENT{with $\Ls^{\mathcal{C}}$ the classifier loss}
    \ENDIF
\ENDFOR

\STATE $\epsilon_{k+1}$ = \textsc{scheduleFunction}($\epsilon_k$, $k$)
\ENDFOR
\STATE \textbf{Return} $\theta_K$, $\phi_K$
\STATE
\STATE
\STATE $\textbf{CollectTrajectories}$($\pi_{\theta}$, $\mathcal{C}_{\phi}$, $\epsilon$, $\sD$, $\tau=0.5$)
\STATE Receive initial observation $\vs_0$
\FOR{$t = 1, 2, ..., T$}
    \STATE $\vm_t \leftarrow \1$
    \IF {$\textsc{random}(0, 1) \leq \epsilon$} \label{alg_line:random_use_classifier}
        \FOR{$a \in \sA$} \label{alg_line:compute_mask}
            \STATE $\evm_{t_a} \leftarrow \mathcal{C}_{\phi}(a, \vs_t) \geq \tau$ \COMMENT{compute the actions mask} 
        \ENDFOR
    \ENDIF
    \STATE Select action $a_t \sim \textsc{softmax}(\vm_t \cdot \mathbf{\pi_{\theta}(\vs_t)})$ \COMMENT{apply the mask} \label{alg_line:apply_mask}
    \STATE Execute action $a_t$ in the environment, observe reward $r_t$ and new state $\vs_{t+1}$
    \STATE $y_t \leftarrow \1_{\vs_{t+1} \neq \vs_t}$ \COMMENT{evaluate whether $a_t$ was applicable} \label{alg_line:applicability}
    \STATE $\sD \leftarrow \sD \cup \{\langle \vs_t, a_t, r_t, \vs_{t+1}, y_t, \vm_t \rangle\}$
\ENDFOR
\end{algorithmic}
\end{algorithm}

\subsection{Learning Inapplicable Actions}

For well understood tasks and environments, such as games where the rules are explicit and unambiguous, the definition of the applicability function is simple and straightforward to encode. However, in cases where the logic driving the (in)applicability of actions is unknown, or even partially understood, it becomes much more challenging to provide valid information to the RL algorithm to mask inapplicable actions. To address this limitation, we propose to learn the applicability function during training by leveraging the data collected by the agent to train the policy. 

Decoupling the inapplicable actions learning from the policy training has the advantage of reducing the problem complexity. 
While the optimality of an action with respect to the total return is observed with a delay, it is possible to directly identify whether an action is applicable in a given state by evaluating the state of the environment before and after it was performed. The reduced problem of identifying the applicability of an action in a given state is thus easier to solve than finding the optimal action.
Learning inapplicable actions also offers a more interpretable way to understand the policy learnt by the RL agent as we can now recognize that some actions are not only sub-optimal in certain states but also inapplicable, providing more information about the learnt policy.

The task of learning the applicability function can in fact be reduced to learning a classifier $\mathcal{C}_{\phi}$, parameterized by $\phi$, that returns the probability for an action $a$ to be applicable in a given state $\vs$. It is possible to train the classifier via supervised learning jointly with the policy and thus provide an increasingly accurate mask of inapplicable actions to the RL algorithm.

We present in \algoref{alg:algorithm} the steps of the algorithm training a classifier jointly with the policy and we detail below the important modifications made to the original algorithm~\citep{Sutton_McAllester_Singh_Mansour_1999}.

\textbf{Inapplicable Actions Masking: }The procedure to collect trajectories is modified to generate the mask using the classifier (\algolineref{alg_line:compute_mask}) and apply it to the policy output (\algolineref{alg_line:apply_mask}).

\textbf{Exploration: }To accommodate the fact that the classifier is learning and is therefore susceptible to produce false negatives (i.e masking out applicable actions) that could be part of an optimal policy, we include an exploration parameter $\epsilon$ driving the frequency at which the mask will be ignored (\algolineref{alg_line:random_use_classifier}). This parameter both helps the classifier collect positive and negative examples to improve its accuracy, but also drives how trusted the classifier can be. The value of $\epsilon$ can vary throughout the training as the classifier improves at identifying applicable actions.

\textbf{Rollout Buffer: }The data buffer collecting the rollouts is extended to include whether or not the action taken was applicable in the state the agent was in. This is computed by comparing the state of the environment before and after the action was performed (\algolineref{alg_line:applicability}). This value will be used as the output the classifier needs to predict. The mask used to select an action is also added to the data buffer, and is used to evaluate the effect of choosing the selected action.

\textbf{Classifier Training: }The classifier $\mathcal{C}_{\phi}$ is trained via supervised learning jointly with the policy $\pi$ over multiple epochs, using the data from the rollout buffer. A key observation is that the policy is learning to act optimally and therefore avoiding taking inapplicable actions, while the classifier also helps filtering out invalid actions. This results in an imbalanced number of positive and negative examples of inapplicable actions in the buffer as the training progress and the agent converges towards an optimal policy. Classifiers are known to perform poorly on imbalanced datasets~\citep{sun2009classification}; to alleviate this issue we re-balance applicable and inapplicable actions samples \algolineref{alg_line:balance_data} using a Weighted Random Sampling approach with the weight inversely proportional to the number of samples for each class.


\subsection{Transfer Learning}

While the search for an optimal policy relies on the reward signal associated with the task at hand, the notion of inapplicable actions depends solely on the environment the agent evolves in. 
%
%
Decoupling the reward signal from the classifier offers the advantage of capturing knowledge about the environment only, without introducing the bias associated with the task the agent is trying to solve. This allows the classifier to be reused to solve different tasks in the same environment in a more efficient manner. Additionally, two different domains with an overlapping action space, are likely to share some properties that make an action inapplicable in a given state. Although, the classifier may not be able to perfectly identify inapplicable actions in a new domain, it already provides some valuable knowledge about the environment that can improve the learning efficiency. The knowledge encapsulated by the classifier can therefore be transferred even across different domains.

\algoref{alg:algorithm} requires little effort to make use of the knowledge previously acquired. By simply passing the parameters of a fitted classifier as input, the algorithm can directly leverage the knowledge acquired about inapplicable actions and mask them away from the policy distribution. To accommodate differences among tasks, \algoref{alg:algorithm} keeps exploring the action space by ignoring the mask generated with the classifier with a random probability. Furthermore, as the classifier continues to be trained along with the policy, it adapts to the new environment and corrects itself to identify inapplicable actions in the new environment.

\section{Experiments}

We present in this section a range of experiments aiming to answer the following research questions. \textbf{Q1:} Does the introduction of human knowledge to mask inapplicable actions help an RL algorithm, such as PPO, to be more sample efficient? \textbf{Q2:} Instead of hand-coding this knowledge, can we learn to identify inapplicable actions while training a policy and use the information acquired to improve the sample efficiency of the RL algorithm? \textbf{Q3:} Can the domain knowledge fed into the algorithm be completed with the knowledge acquired by the inapplicable actions classifier to further improve the performance of the algorithm? \textbf{Q4:} Is it possible to share the knowledge acquired across tasks and domains to help improve the training efficiency of new policies?

\begin{figure}[h!]
\centering
\captionsetup{justification=centering}
\captionsetup[sub]{font=scriptsize}
\hspace{\fill}
\begin{subfigure}[t]{0.075\textwidth}\centering
    \includegraphics[width=\textwidth]{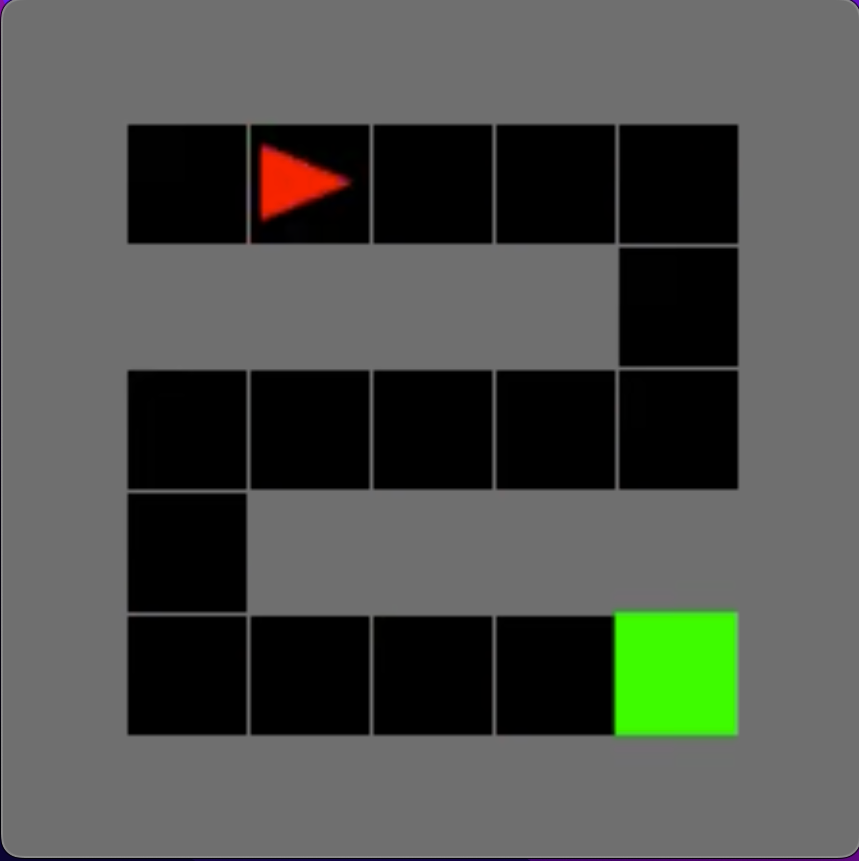}
    \caption{Maze}
    \label{fig:maze_environment}
\end{subfigure}
\hfill
\begin{subfigure}[t]{0.075\textwidth}\centering
    \includegraphics[width=\textwidth]{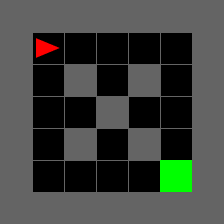}
    \caption{X-Island (1)}
    \label{fig:xisland_environment}
\end{subfigure}
\hfill
\begin{subfigure}[t]{0.075\textwidth}\centering
    \includegraphics[width=\textwidth]{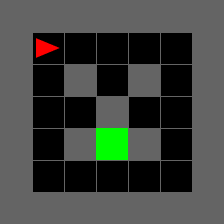}
    \caption{X-Island (2)}
    \label{fig:xislandbis_environment}
\end{subfigure}
\hfill
\begin{subfigure}[t]{0.075\textwidth}\centering
    \includegraphics[width=\textwidth]{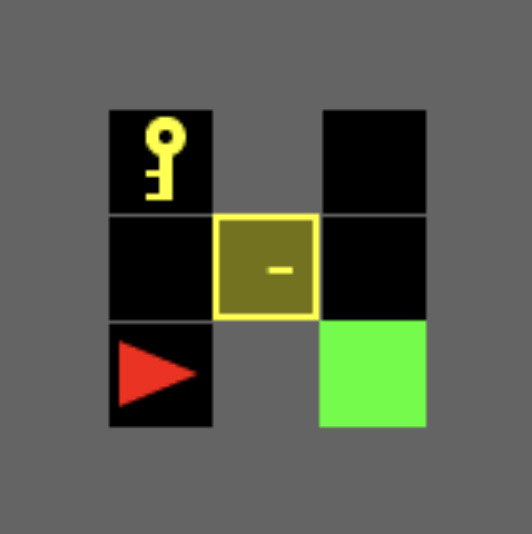}
    \caption{Key \& Door (1)}
    \label{fig:doorkey_environment}
\end{subfigure}
\hfill
\begin{subfigure}[t]{0.075\textwidth}\centering
    \includegraphics[width=\textwidth]{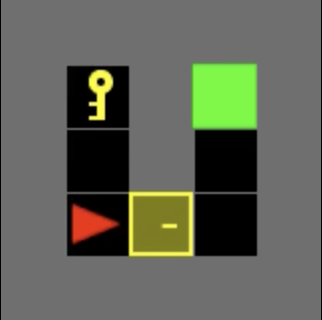}
    \caption{Key \& Door (2)}
    \label{fig:doorkeybis_environment}
\end{subfigure}
\hspace{\fill}
\caption{Different tasks in the Maze and Key \& door domains.}
\label{fig:environment}
\end{figure}

\begin{figure*}[h]
\centering
\captionsetup{justification=centering}
\captionsetup[sub]{font=scriptsize}
\captionsetup[sub]{justification=centering}

\begin{subfigure}[t]{0.45\textwidth}\centering
    \includegraphics[width=0.5\textwidth]{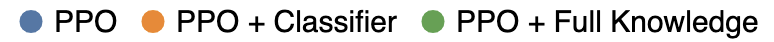}
\end{subfigure}

\begin{subfigure}[t]{0.3\textwidth}\centering
    \includegraphics[width=\textwidth]{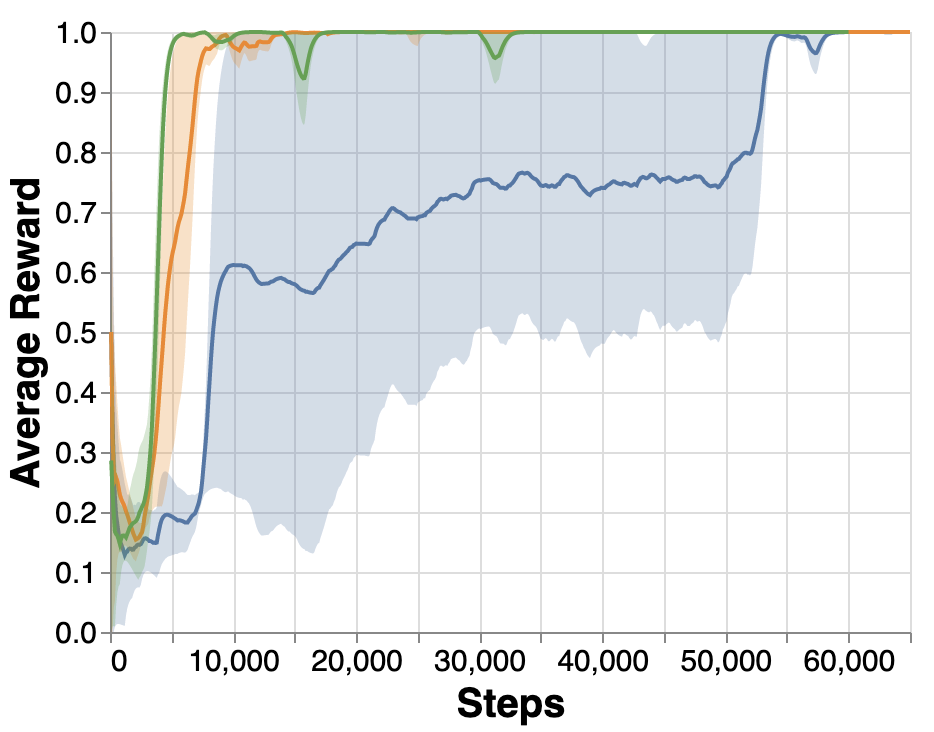}
    \caption{Average reward per episode}
    \label{fig:maze_reward_ppo_classifier}
\end{subfigure}
\begin{subfigure}[t]{0.3\textwidth}\centering
    \includegraphics[width=\textwidth]{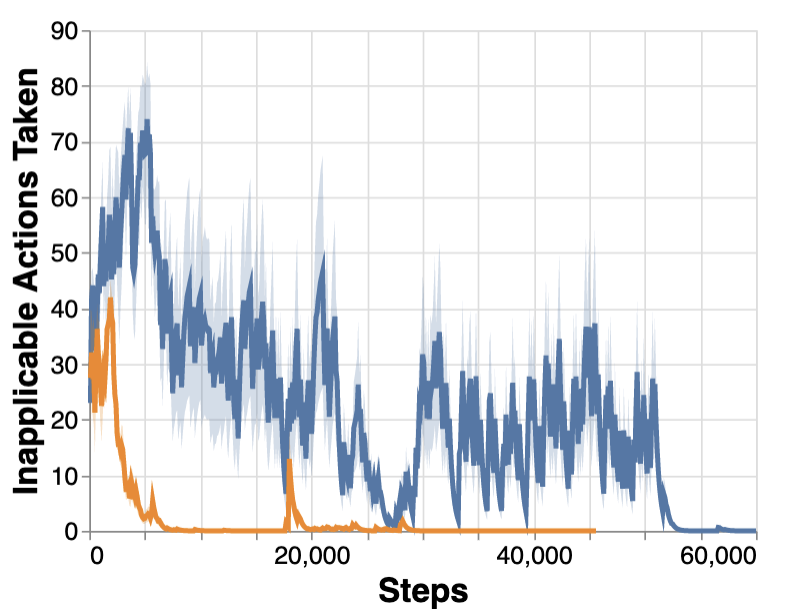}
    \caption{Average number of inapplicable actions taken per episode}
    \label{fig:maze_inapplicable_actions_ppo_classifier}
\end{subfigure}
\begin{subfigure}[t]{0.3\textwidth}\centering
    \includegraphics[width=\textwidth]{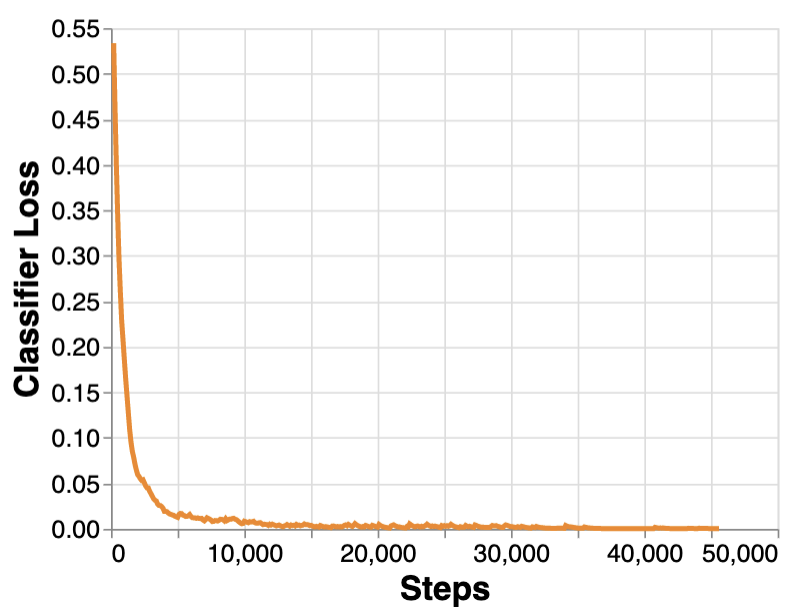}
    \caption{Classifier loss}
    \label{fig:maze_classifier_loss_ppo_classifier}
\end{subfigure}

\caption{Learning the inapplicable actions mask in the Maze environment.}
\label{fig:maze_ppo_classifier}
\end{figure*}\

The set of experiments presented are run on $5$ tasks from $2$ different domains shown in \figref{fig:environment} where an agent (in red) starting at a random position moves in a gridworld-like environment to reach a fixed goal cell (in green). The tasks are designed so that the agent will only be rewarded once it reaches the goal, and the reward received will be inversely proportional to the number of steps taken, making the agent learn how to reach the goal in the fewest possible steps. In these experiments, states are given as images as opposed to most work on automated planning that assumes a propositional representation. At each time step, the agent receives the full image of the environment to make a decision on the action to perform. In all the environments the agent can go \texttt{up}, \texttt{right}, \texttt{down}, \texttt{left}. In the Key \& Door domain, it can also decide to \texttt{pickup} a key and \texttt{open} a door. The agent must pickup the key before opening the door. But, once the door is open, it remains so and does not require to be opened again. 

We present the average reward across 5 runs normalized by the maximum reward an agent can receive, the number of inapplicable actions the agent took and the classifier loss. For this set of experiments, the environment being deterministic, we use the identity function to compute the distance between two states ($d(\vs_{t+1}, \vs) = 1 - \1_{\vs_{t+1} = \vs}$).

The experiments configuration and the code are provided in the supplementary materials.

\subsection{Policy Learning with Domain Knowledge}

\noindent\textbf{Setup:}
%
%
We manually encoded (perfect) domain knowledge in the X-Island and Key \& Door tasks depicted in Figures~\ref{fig:xisland_environment} and~\ref{fig:doorkey_environment}, respectively. Then, we ran experiments to compare the learning curves of an agent with full versus zero knowledge of the inapplicable actions.

\noindent\textbf{Objectives:} These experiments aim at providing some evidence about the practical impact of the approach of introducing inapplicable actions knowledge, as well as assessing its potential gains. 

\begin{figure}[H]
\centering
\captionsetup{justification=centering}
\captionsetup[sub]{font=scriptsize}
\begin{subfigure}[t]{0.3\textwidth}\centering
    \includegraphics[width=\textwidth]{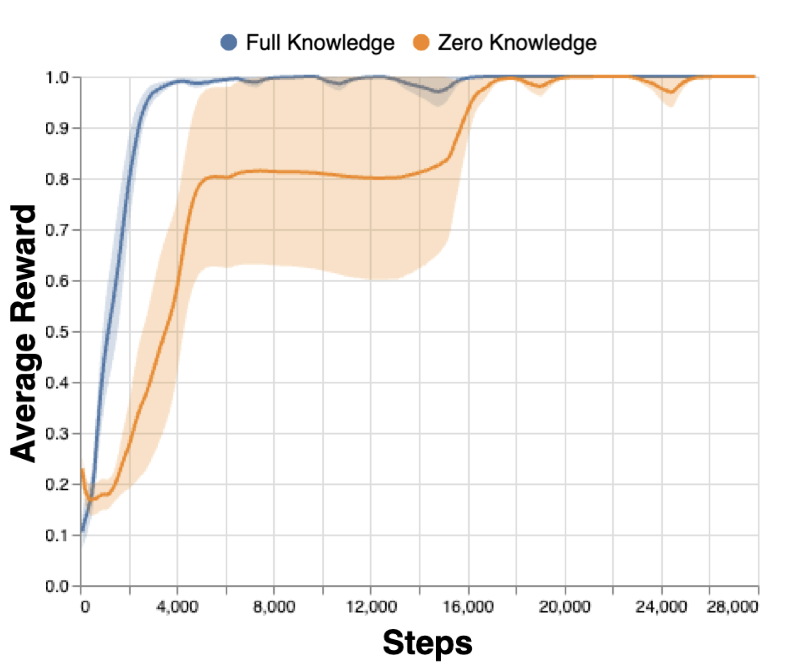}
    \caption{Full vs Zero Knowledge in the X-Island task.}
    \label{fig:X_island_full_zero}
\end{subfigure}
\begin{subfigure}[t]{0.3\textwidth}\centering
    \includegraphics[width=\textwidth]{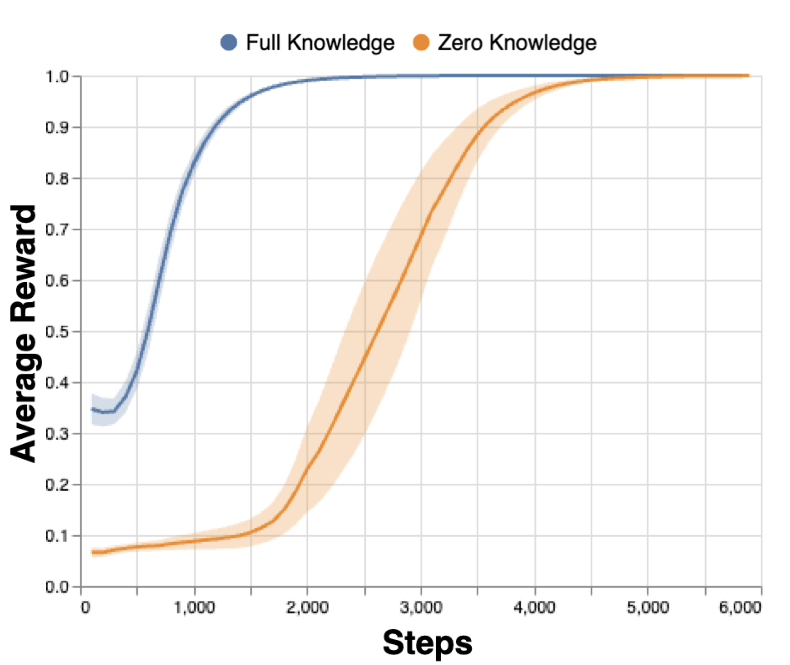}
    \caption{Full vs Zero Knowledge in the Key \& Door (1) task.}
    \label{fig:door_key_full_zero}
\end{subfigure}
\caption{Average reward per episode in different tasks with different 
 levels of inapplicable actions knowledge.}
\label{fig:knowledge_reward}
\end{figure}

\noindent\textbf{Results:} The results of these experiments are shown in Figures \ref{fig:X_island_full_zero} and \ref{fig:door_key_full_zero} for the X-Island (1) and Key \& Door (1) tasks, respectively.
As expected, the agent converges faster in both tasks when it has perfect knowledge about the inapplicable actions at each state (blue curves), since it does not need to waste effort exploring useless state-action pairs.
Although observed in both domains, the performance gain is bigger in the Key \& Door domain.
In this case, the inapplicable action knowledge is able to prune $74\%$ of all the possible state-action pairs, while it is \emph{just} pruning $36\%$ of the pairs in the X-Island domain.

\subsection{Policy Training with Inapplicable Actions Learning} \label{sec:exp_classifier_from_scratch}

\noindent\textbf{Setup:} To evaluate the impact of learning inapplicable actions during policy training, we run  \algoref{alg:algorithm} in the $3$ environments presented in \figref{fig:environment} with $5$ different seeds and evaluate the sample efficiency gain against the original PPO algorithm. We use an exploration parameter $\epsilon=0.5$, corresponding to a use of the mask generated by the classifier half of the time allowing the classifier to be exposed to both positive and negative examples of applicability. See \appendixref{sec:neural_network_architectures} for the architecture of the NN used for the classifier.

\noindent\textbf{Objectives:} These experiments consider the situation where no knowledge about inapplicable actions is available at start. We wish to evaluate if learning an estimate of the applicability function $C$ while training the policy is feasible, and whether it can be used directly by the RL algorithm to avoid exploring irrelevant actions.

\noindent\textbf{Results:} The classifier (orange curve) is able to rapidly learn to identify inapplicable actions as suggested by the classifier loss curve (\figref{fig:maze_classifier_loss_ppo_classifier}). As we can see in \figref{fig:maze_inapplicable_actions_ppo_classifier}, the number of inapplicable actions taken by the agent reduces faster than the PPO baseline throughout the training. Hence, the agent avoids exploring inapplicable actions thanks to the classifier and the quality of the mask generated. As a result, this algorithm is more sample efficient than the PPO baseline (\figref{fig:maze_reward_ppo_classifier}). Similar results are observed for the X-Island and the Key \& Door environments (\appendixref{sec:inapplicable_actions_learning}). The more constrained the environment, the bigger the performance improvement will be when we introduce the inapplicable actions learning component. In the Maze domain, the agent has in most cases only 50\% of its action space applicable which explains the significant performance improvement.

\begin{figure}[H]
\centering
\captionsetup{justification=centering}
\captionsetup[sub]{font=scriptsize}
\captionsetup[sub]{justification=centering}
\hspace{\fill}
\begin{subfigure}[t]{0.1\textwidth}\centering
    \includegraphics[width=\textwidth]{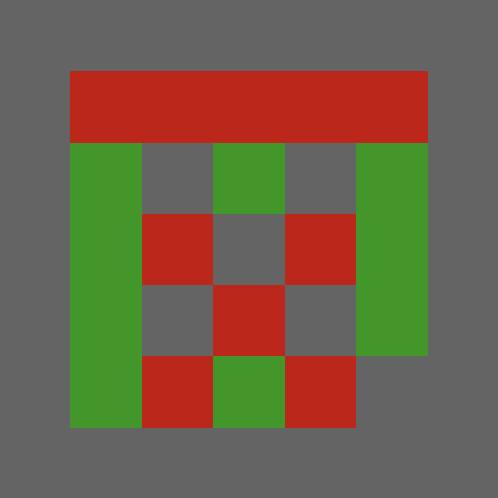}
    \caption{Action \texttt{up}}
    \label{fig:xisland_classifier_up}
\end{subfigure}
\hfill
\begin{subfigure}[t]{0.1\textwidth}\centering
    \includegraphics[width=\textwidth]{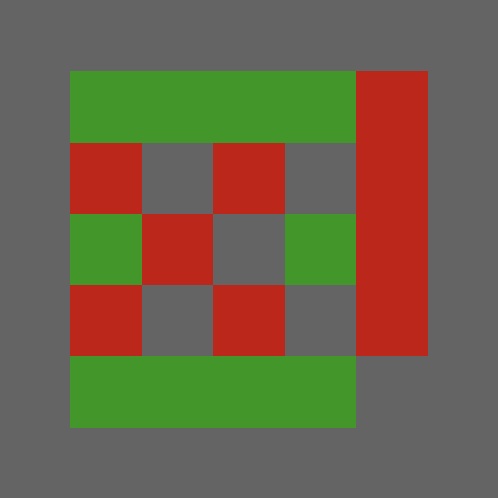}
    \caption{Action \texttt{right}}
    \label{fig:xisland_classifier_right}
\end{subfigure}
\hfill
\begin{subfigure}[t]{0.1\textwidth}\centering
    \includegraphics[width=\textwidth]{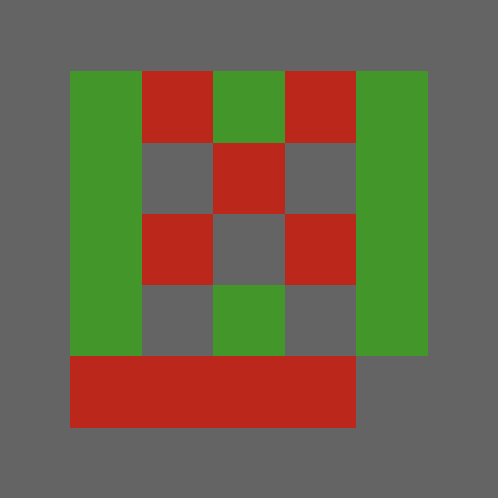}
    \caption{Action \texttt{down}}
    \label{fig:xisland_classifier_down}
\end{subfigure}
\hfill
\begin{subfigure}[t]{0.1\textwidth}\centering
    \includegraphics[width=\textwidth]{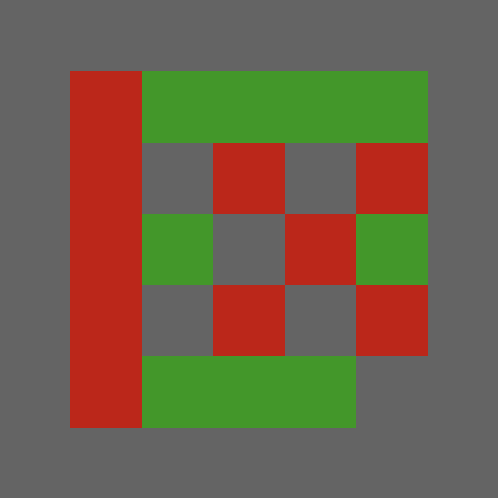}
    \caption{Action \texttt{left}}
    \label{fig:xisland_classifier_left}
\end{subfigure}
\hspace{\fill}

\caption{Classifier output learnt in the X-Island (1) environment}
\label{fig:classifier_analysis}
\end{figure}

\subsubsection{Classifier Analysis}
\label{sec:exp_classifier_analysis}
\ \\

\noindent\textbf{Setup:} We analyze the trained classifier for each of the environments (see also \appendixref{sec:classifier_analysis_additional}) and we present in \figref{fig:classifier_analysis} the predictions the classifier makes on the applicability of each of the actions the agent can take in a given state. A red (respectively green) cell indicates that the classifier has predicted that the action is inapplicable (respectively applicable) when the agent stands on that cell.

\noindent\textbf{Objectives:} Understand what the classifier has learnt and make sure that the knowledge it acquired in the gridworld environment concurs with the common sense of treating actions that go into a wall as inapplicable.

\noindent\textbf{Results:} We see that the classifier is able to perfectly learn inapplicable actions at each given state. For instance, all the cells where the agent has a wall on its left are red in \figref{fig:xisland_classifier_left}. This confirms the hypothesis that the classifier provides valuable information about the environment to the RL algorithm to mask out inapplicable actions.
We can easily see graphically that the results would be the same had we trained/tested on bigger sized tasks.

\begin{figure}[h!]
    \centering
    \captionsetup{justification=centering}
    \includegraphics[width=0.3\textwidth]{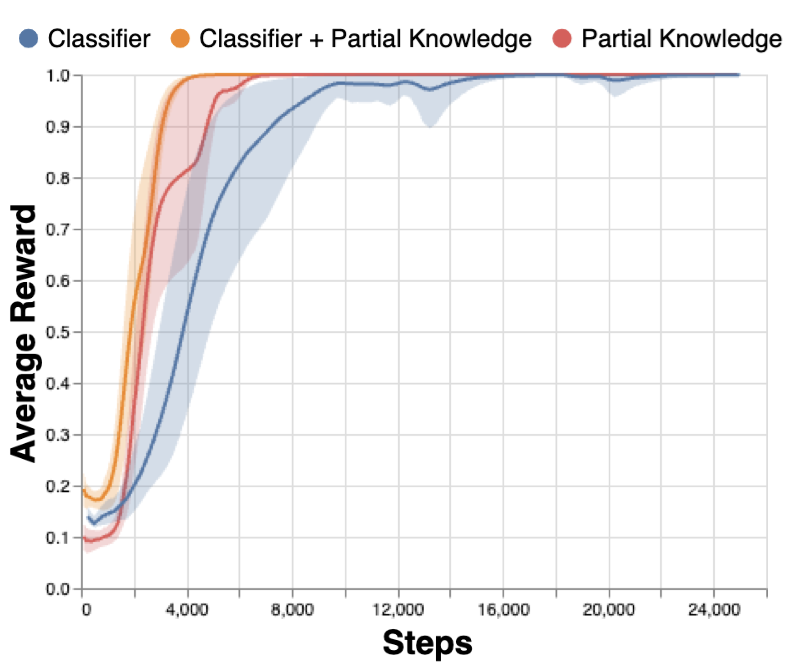}
    \caption{Average reward per episode in the Key \& Door (1) task  with different ways of acquiring the inapplicable actions knowledge.}
    \label{fig:dk_scratch_partial_classifier}
\end{figure}

\begin{figure*}[h]
\centering
\captionsetup{justification=centering}

\hspace{\fill}
\begin{subfigure}[t]{0.45\textwidth}\centering
    \includegraphics[width=\textwidth]{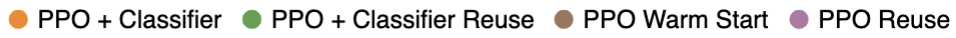}
\end{subfigure}
\hspace{\fill}

\hspace{\fill}
\begin{subfigure}[t]{0.3\textwidth}\centering
    \includegraphics[width=\textwidth]{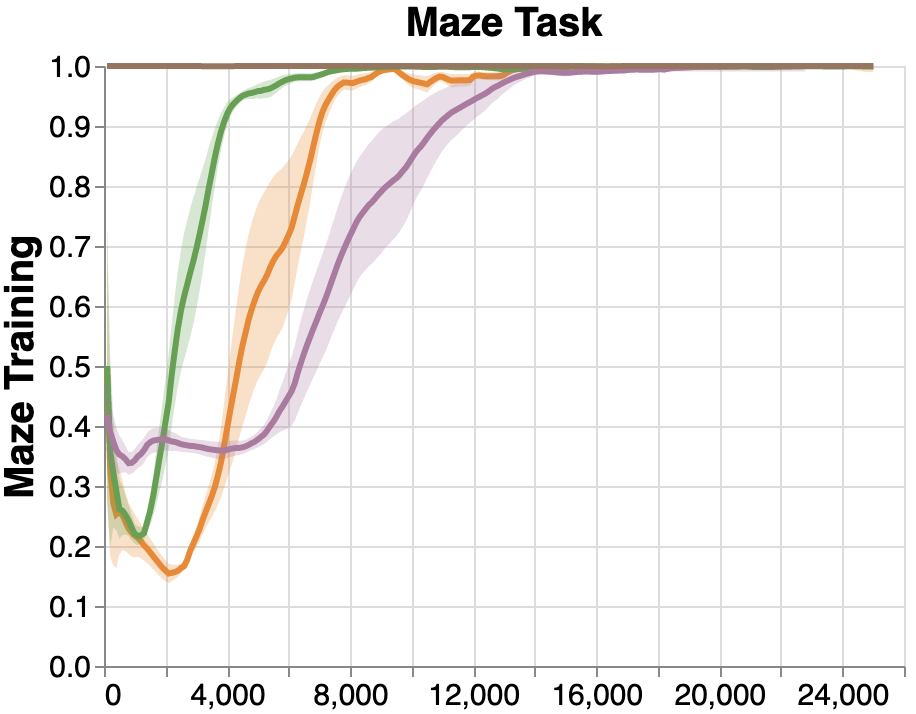}
\end{subfigure}
\begin{subfigure}[t]{0.3\textwidth}\centering
    \includegraphics[width=\textwidth]{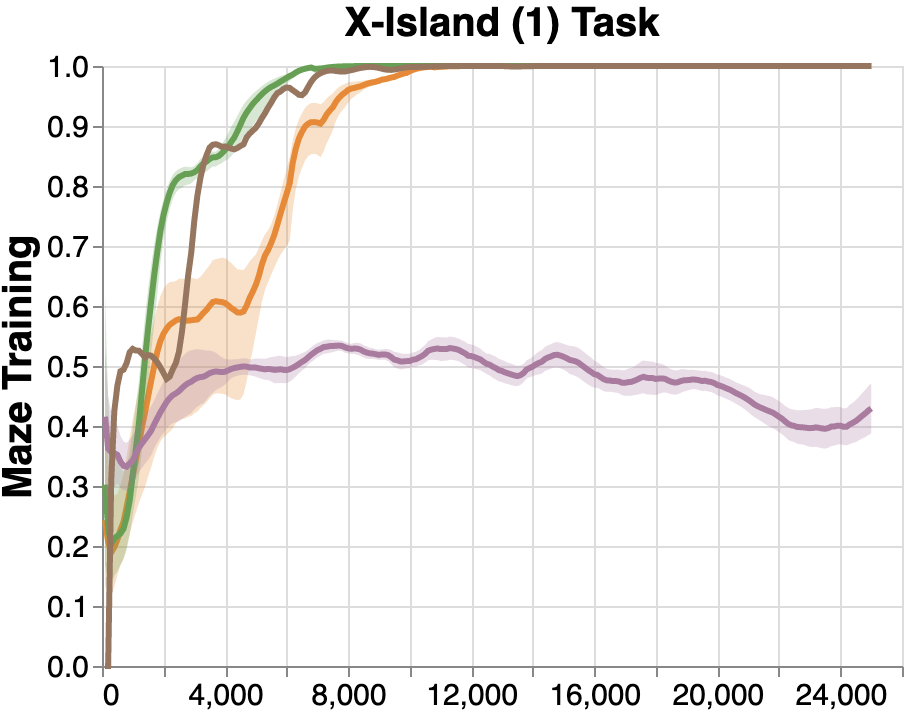}
\end{subfigure}
\begin{subfigure}[t]{0.3\textwidth}\centering
    \includegraphics[width=\textwidth]{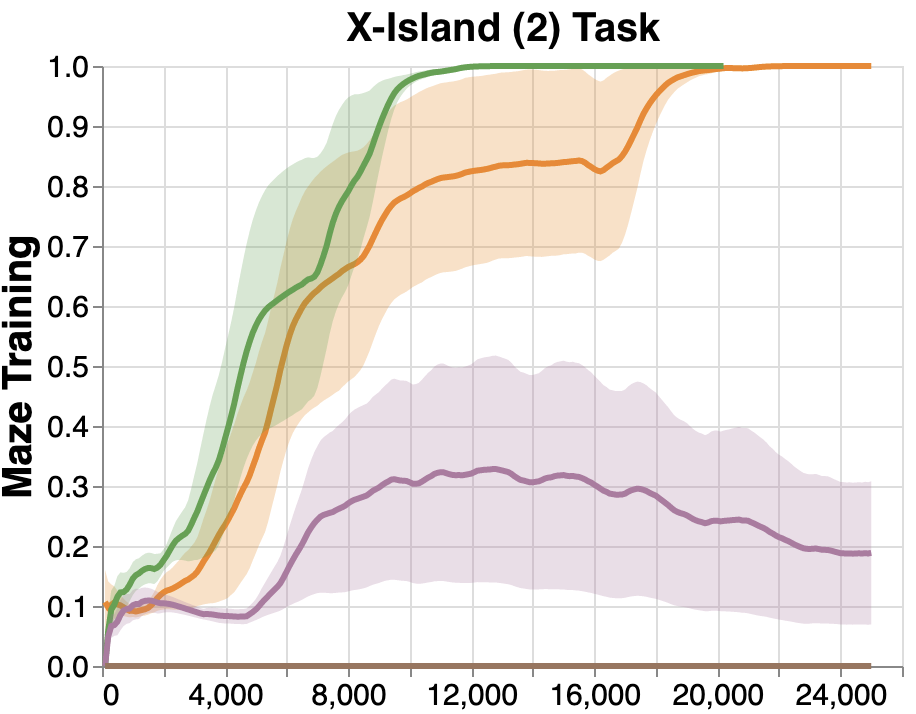}
\end{subfigure}
\hspace{\fill}

\hspace{\fill}
\begin{subfigure}[t]{0.3\textwidth}\centering
    \includegraphics[width=\textwidth]{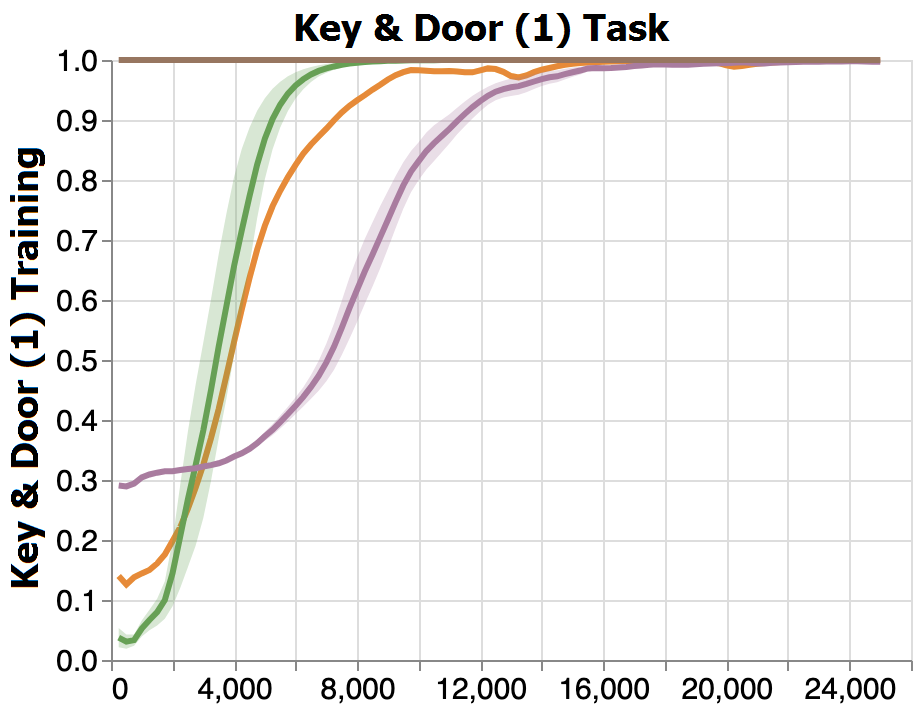}
\end{subfigure}
\begin{subfigure}[t]{0.3\textwidth}\centering
    \includegraphics[width=\textwidth]{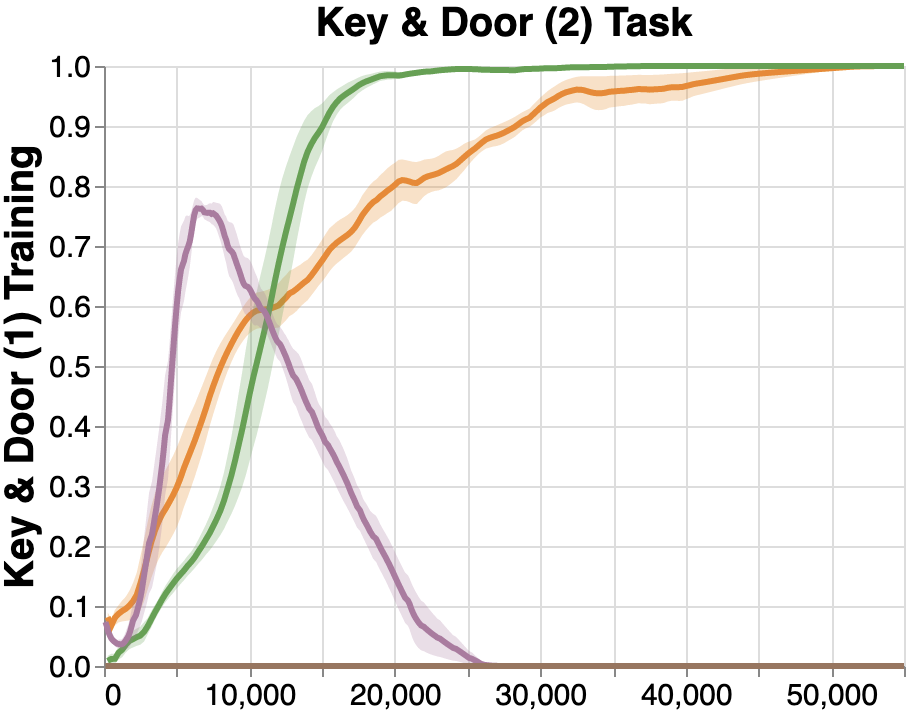}
\end{subfigure}
\hspace{\fill}

\caption{Average reward for different techniques of knowledge transfer reusing the knowledge acquired to solve different tasks. The knowledge is acquired in the environment specified as the y-axis title and reused in the environment specified in the chart title.}
\label{fig:maze_classifier_reuse}
\end{figure*}

\subsubsection{Inapplicable Actions Learning with Partial Domain Knowledge}\hspace*{\fill} \\

\noindent\textbf{Setup:} We specify a subset of the applicability function $C$ in the Key \& Door (1) and X-Island (1) environments.
In Key \& Door we only provide information about the \texttt{up}, \texttt{down}, \texttt{right} and \texttt{left} actions.
In X-Island (1) we only provide information about \texttt{up} and \texttt{down}.

\noindent\textbf{Objectives:} Show how the classifier can learn the missing knowledge about the applicability of some actions.

\noindent\textbf{Results:} The results of this experiment in the Key \& Door environment are shown in \figref{fig:dk_scratch_partial_classifier} (see also \appendixref{sec:partial_plus_classifier})
As we can see, jointly using partial knowledge and the classifier speeds up the learning both over learning the classifier from scratch and just having partial knowledge.

\subsection{Knowledge Transfer Across Tasks} \label{sec:knowledge_transfer_tasks}

\noindent\textbf{Setup:} We solve again the tasks presented in \figref{fig:environment}. This time we provide the algorithm with the trained classifiers obtained in the same domain but while solving a different task. We reduce the exploration parameter $\epsilon$ to $0.25$ as we have more confidence in the trained classifier. We compare our technique with an implementation of Policy Reuse~\citep{Fernandez_Garcia_Veloso_2010} using an exploration parameter of $0.25$ and a warm start technique initializing the policy with the weights of a pre-trained policy.

\noindent\textbf{Objectives:} Check if the knowledge encapsulated in the inapplicable actions classifier can be transferred across different tasks within the same domains to help improve the learning efficiency of new policies.

\noindent\textbf{Results:} We show in \figref{fig:maze_classifier_reuse} the impact of reusing a classifier trained in the Maze Environment when solving different tasks in the same domain (X-Island (1), X-Island (2)). We also present the results of reusing a classifier trained in the Key \& Door (1) Environment to solve the Key \& Door (2) task. The transfer of the classifier yields better performance than learning the classifier from scratch. We also note that our approach yields better results than PPO reuse or a naive warm start approach which tend to overfit to the goal the policy was trained for. The knowledge acquired by the classifier is indeed task agnostic and can therefore be learnt in simpler tasks to help solving more complex problems.

\subsection{Knowledge Transfer Across Domains}

\begin{figure}[h!]
\centering
\captionsetup{justification=centering}

\hspace{\fill}
\begin{subfigure}[t]{0.45\textwidth}\centering
    \includegraphics[width=\textwidth]{figures/transfer/legend.png}
\end{subfigure}
\hspace{\fill}

\begin{subfigure}[t]{0.3\textwidth}\centering
    \includegraphics[width=\textwidth]{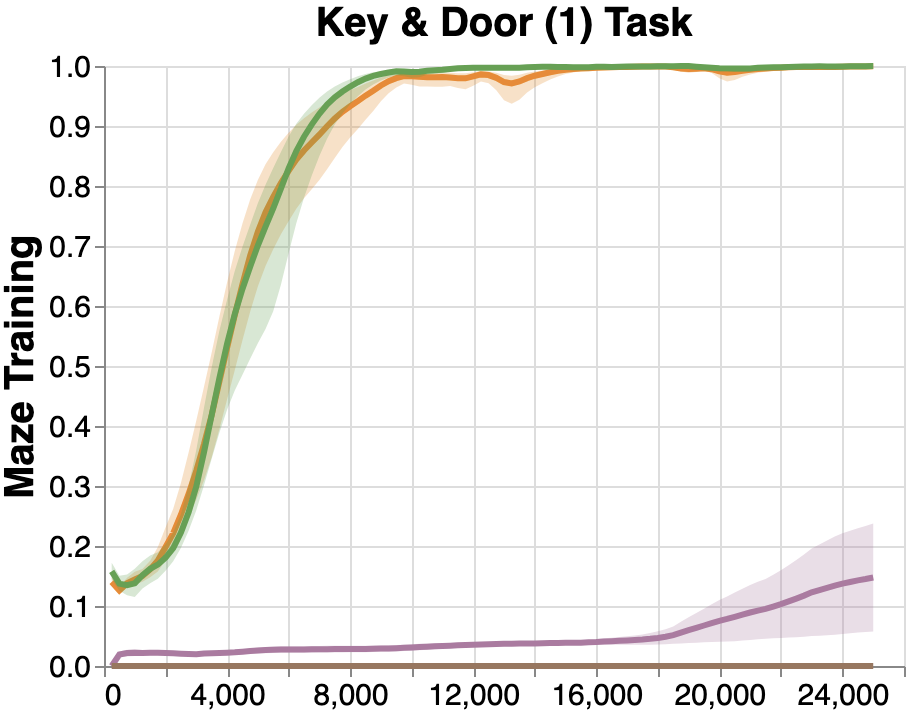}
\end{subfigure}
\caption{Average reward for different techniques of knowledge transfer reusing the knowledge acquired in the Maze environment to solve the Key \& Door task.}
\label{fig:domain_classifier_reuse_maze}
\end{figure}

\noindent\textbf{Setup:} We now evaluate whether the knowledge acquired in one domain can be transferred to another. We reuse the classifier trained in the Maze environment to solve the Key \& Door task and vice versa. These two environments share domain constraints such as the walls. However, the Key \& Door environment is more complex with the need to have the key to be able to open the door. Like in \secref{sec:knowledge_transfer_tasks}, we use Policy Reuse~\citep{Fernandez_Garcia_Veloso_2010} and a warm start as a baseline to evaluate the performance of our approach. We also use an exploration value of $\epsilon=0.25$.

\noindent\textbf{Objectives:} Evaluate if the knowledge encapsulated by the trained inapplicable action classifier can be used to improve the learning efficiency of a new task in a different domain.

\begin{figure}[h!]
\centering
\captionsetup{justification=centering}
\hspace{\fill}
\begin{subfigure}[t]{0.45\textwidth}\centering
    \includegraphics[width=\textwidth]{figures/transfer/legend.png}
\end{subfigure}
\hspace{\fill}

\begin{subfigure}[t]{0.3\textwidth}\centering
    \includegraphics[width=\textwidth]{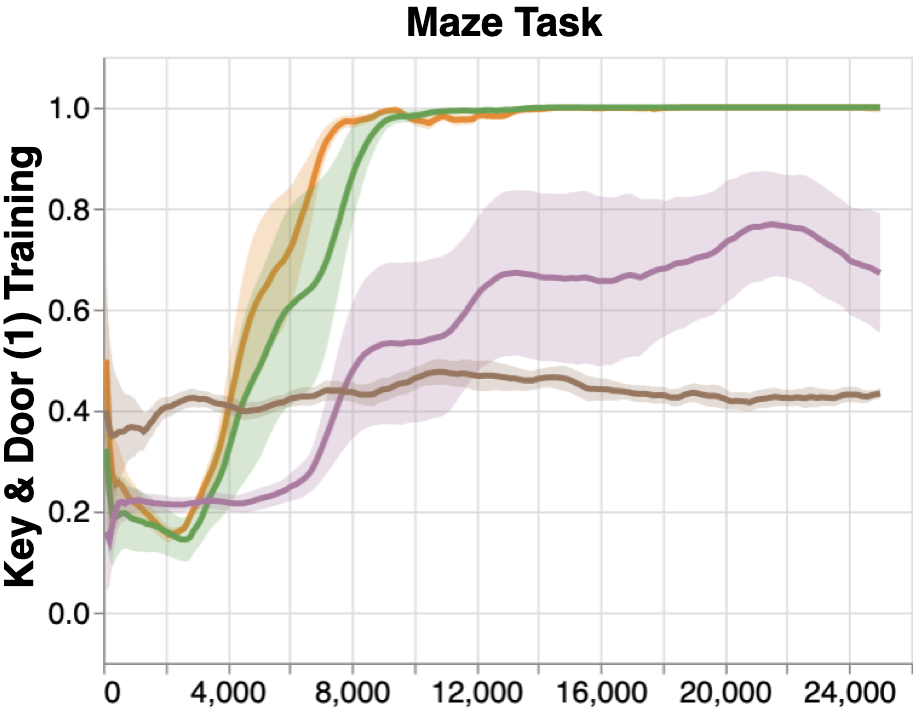}
\end{subfigure}

\caption{Average reward for different techniques of knowledge transfer reusing the knowledge acquired in the Key \& Door environment to solve the Maze task.}
\label{fig:domain_classifier_reuse_doorkey}
\end{figure}

\noindent\textbf{Results:} Figures~\ref{fig:domain_classifier_reuse_maze} and~\ref{fig:domain_classifier_reuse_doorkey} present the results of transferring the knowledge acquired in one domain to another. We see in \textit{green}, the performance of our approach when we reuse the inapplicable actions classifier and in \textit{orange}, the effect of learning the classifier from scratch. 
We see that in both case, providing a trained inapplicable actions classifier helps the algorithm reach an optimal policy faster.
Policy Reuse and Warmstart being biased towards the task, behave poorly in a new domain. This highlights the strength of learning in a task agnostic way.
The Key \& Door task being arguably more complex than the Maze one, we show here how our approach can be used to train a classifier in a simpler domain and transfer the knowledge acquired to a more complex one to help improve the learning efficiency.

\section{Related Work}

The use of actions masking has been used by the RL community to improve the sample efficiency of RL algorithms and has contributed to some of the most impressive breakthroughs in the field~\citep{Vinyals_2017, kool2018attention}. Despite being used, the effect of action masking had not been thoroughly analyzed until recently with \citet{huang2022closer}, where the theoretical soundness of the approach is shown when used with Policy Gradient algorithms.
%
%

Only a few papers tackle the problem of learning the masking function.
\citet{Even-Dar_Mannor_Mansour_2006} appears to be the first to look at actions eliminations in the context of RL by using the confidence interval around the Q-function to avoid actions that are sub-optimal with high probability. Later,~\citet{zahavy2018learn} present a new algorithm combining DQN and an Action Elimination Network (AEN) learning to mask out sub-optimal actions. 
In both works the focus is put on sub-optimal actions and is therefore biased by the task to solve.
Our approach on the other hand focuses on learning a particular feature of the environment only. This specificity allows us to transfer the knowledge gained in one environment to another.

To reuse previously acquired knowledge, ~\citet{pmlr-v80-barreto18a} associate the Successor Representations (SR) framework, decoupling the state features and the reward distribution to estimate the value function, with the General Policy Improvement framework (GPI). This technique assumes, however, that the tasks only differ by their reward distribution.
Close to our approach, the Policy Reuse algorithm ~\citep{Fernandez_Garcia_Veloso_2010} uses expert policies previously trained to help orient the learning agent towards an optimal policy. 

%
%
Recent works from the automated planning research community try to learn the symbolic model capturing the applicability of actions at each state either from latent spaces~\citep{DBLP:conf/aaai/AsaiF18} or from the structure of the state space~\citep{DBLP:conf/ecai/BonetG20}.
Our approach differs from these works in two main ways.
First, we are generating the image pairs needed by Asai and Fukunaga (\citeyear{DBLP:conf/aaai/AsaiF18}) by allowing the agent to interact with the environment and learn a policy, as opposed to assuming image pairs are provided a priori.
Second, while most of these works focus on learning a planning model to solve several tasks in the same domain, our learned inapplicable actions classifier can be reused to solve tasks not only in the same domain but also in domains that share certain similarities with the original one.


%

\section{Conclusion}

Inspired by the Automated Planning literature and the use of STRIPS to define an action model, we formalize in this work a systematic way to provide and make use of the action \textit{precondition} in the RL algorithm. It helps improve the sample efficiency of the algorithms by limiting the exploration to actions that are relevant for finding an optimal policy.
We also proposed a technique to learn this partial action model, when it is not known a priori. We train jointly with the policy a classifier predicting that an action is inapplicable in a given state. The information captured by this new component can directly be used to mask out inapplicable actions, leading to more sample efficient RL algorithms.
The partial action model learnt encapsulates knowledge about the environment that can be transferred to other tasks and other similar domains, providing an even greater performance gain. 

The transferability of the knowledge acquired is, however, subject to certain conditions. For one, the two domains must share a subset of their action space as the classifier depends on the action to predict the applicability of an action. The two domains must also share some common features that make a given action inapplicable for the transfer to be efficient. An interesting research avenue could be to first understand how different domains can be to be able to transfer knowledge.
While we have shown that our approach works in deterministic fully observable settings, we leave as future work the extension of this approach to more complex domains with stochasticity and only partial observability.



\subsubsection*{Disclaimer}
This paper was prepared for informational purposes by the Artificial Intelligence Research group of JPMorgan Chase \& Co and its affiliates (“J.P. Morgan”), and is not a product of the Research Department of J.P. Morgan. J.P. Morgan makes no representation and warranty whatsoever and disclaims all liability, for the completeness, accuracy or reliability of the information contained herein. This document is not intended as investment research or investment advice, or a recommendation, offer or solicitation for the purchase or sale of any security, financial instrument, financial product or service, or to be used in any way for evaluating the merits of participating in any transaction, and shall not constitute a solicitation under any jurisdiction or to any person, if such solicitation under such jurisdiction or to such person would be unlawful. © 2022 JPMorgan Chase \& Co. All rights reserved.

\newpage

\bibliography{references}

\newpage
\appendix
\section{Experiments configuration} \label{sec:experiments_configuration}

All the experiments were run on a \textit{r6i.8xlarge} EC2 instance. The results presented are the mean value over 5 different seeds using the standard error to construct the confidence interval. For all the experiments presented, the policy was trained until convergence or stopped after $100,000$ steps. The implementation of the PPO algorithm from~\citep{stable-baselines3} was used as a baseline, and modified to implement the \algoref{alg:algorithm} introducing the inapplicable actions masking components. 

The code for the Open AI Gym environments used and the RL algorithms including \algoref{alg:algorithm} is provided in the supplementary materials.

For the experiments learning inapplicable actions, the classifier is trained jointly with the policy, using the same batch size ($64$), the same number of epochs ($10$) per training iteration and with an Adam optimizer using a learning rate of $3\cdot10^{-4}$ for the \textit{Inapplicable Actions Learning} experiments and $1\cdot10^{-4}$ for the \textit{Knowledge Transfer} experiments. The classifier used is a neural network composed of three elements. The first element called the \texttt{ObservationsExtractor} is a Convolutional Neural Network (CNN) in charge of extracting features from the image representing the current state of the system. The second component, \texttt{ActionsExtractor}, simply one-hot encodes the action that needs to be evaluated. Finally, the classifier is a binary classifier implemented with a Multi-Layer Perceptron using ReLU and Dropout layers, that takes as an input the concatenation of the vectors from the two extractor components and output the logit of the action being applicable in the given state. The policy neural network uses a features extractor network generating features in a latent space that will then be fed into the policy and the value networks of the PPO algorithm~\citep{stable-baselines3}.

The full architecture details of the neural networks are provided in \appendixref{sec:neural_network_architectures}.

\begin{table*}[h!]
\caption{Neural Network architectures}
\label{tab:nn_observations_extractor}
\begin{center}
\begin{tabular}{l l || l}
\hline
\multicolumn{1}{c}{\textbf{Observations Extractor}} & \multicolumn{1}{c||}{\textbf{Classifier}} & \multicolumn{1}{c}{\textbf{Policy Features Extractor}} \\
\hline
BatchNorm2d(N) & BatchNorm1d(M) & Conv2d(N, 32, kernel\_size=8, stride=4) \\
Conv2d(N, 32, kernel\_size=8, stride=4) & Linear(M, 256) & ReLU() \\
ReLU() & ReLU() & Conv2d(32, 64, kernel\_size=4, stride=2) \\
Conv2d(32, 64, kernel\_size=4, stride=2) & Dropout(0.3) &  ReLU()\\
ReLU() & BatchNorm1d(256) & Conv2d(64, 64, kernel\_size=3, stride=1) \\
Conv2d(64, 64, kernel\_size=3, stride=1) & Linear(256, 96) & ReLU() \\
ReLU() & ReLU() & Flatten() \\
Flatten() & Dropout(0.3) \\
& BatchNorm1d(96) & \\
& Linear(96, 1) & \\
\hline
\end{tabular}
\end{center}
\end{table*}

\section{Neural network architectures} \label{sec:neural_network_architectures}

All the neural networks' weights are initialized using the \textit{Xavier} initialization, also known as \textit{Glorot} initialization, associated to the uniform distribution. We use the $\texttt{Pytorch}$ framework~\citep{NEURIPS2019_9015} to implement all the neural network described in this paper.

\section{Domain knowledge}

\subsection{Different levels of domain knowledge}

\begin{figure}[ht]
\centering
\captionsetup{justification=centering}
\includegraphics[width=0.3\textwidth]{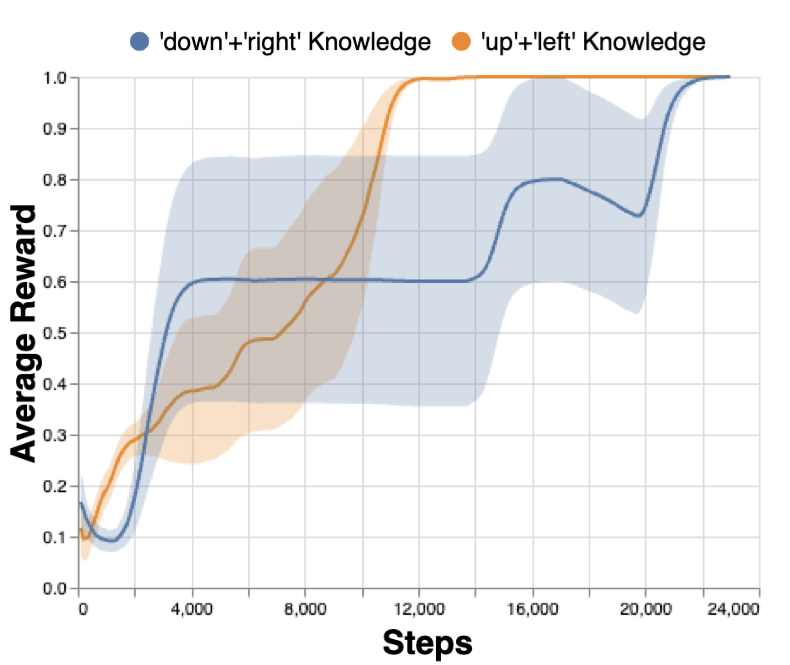}
\caption{Importance of the level of inapplicable actions knowledge provided to the algorithm}
\label{fig:X_island_upleft_downright}
\end{figure}

The performance gain provided by knowing actions (in)applicability not only depends on the task but also on the specific actions we have information from.
Consider the X-Island (1) task depicted in \figref{fig:xisland_environment}.
In this case, the optimal policy involves the agent taking a number of down and right actions.
Therefore, having knowledge about when the agent cannot move \texttt{up} or \texttt{left} (actions that the agent does not need to execute) should accelerate the learning more than having information about the move \texttt{down} and \texttt{right} actions (actions that the agent will need to execute).
This hypothesis is confirmed by the results in \figref{fig:X_island_upleft_downright}, where we can observe that the agent converges faster when it has access to the \texttt{up} and \texttt{left} actions knowledge than when it has access to the \texttt{down} and \texttt{right} knowledge.

\section{Inapplicable actions learning}
\label{sec:inapplicable_actions_learning}

\begin{figure*}
\centering
\captionsetup{justification=centering}
\captionsetup[sub]{font=scriptsize}
\hspace{\fill}
\begin{subfigure}[t]{0.45\textwidth}\centering
    \includegraphics[width=0.5\textwidth]{figures/legend_ppo_vs_classifier_2.png}
\end{subfigure}
\hspace{\fill}

\begin{subfigure}[t]{0.3\textwidth}\centering
    \includegraphics[width=\textwidth]{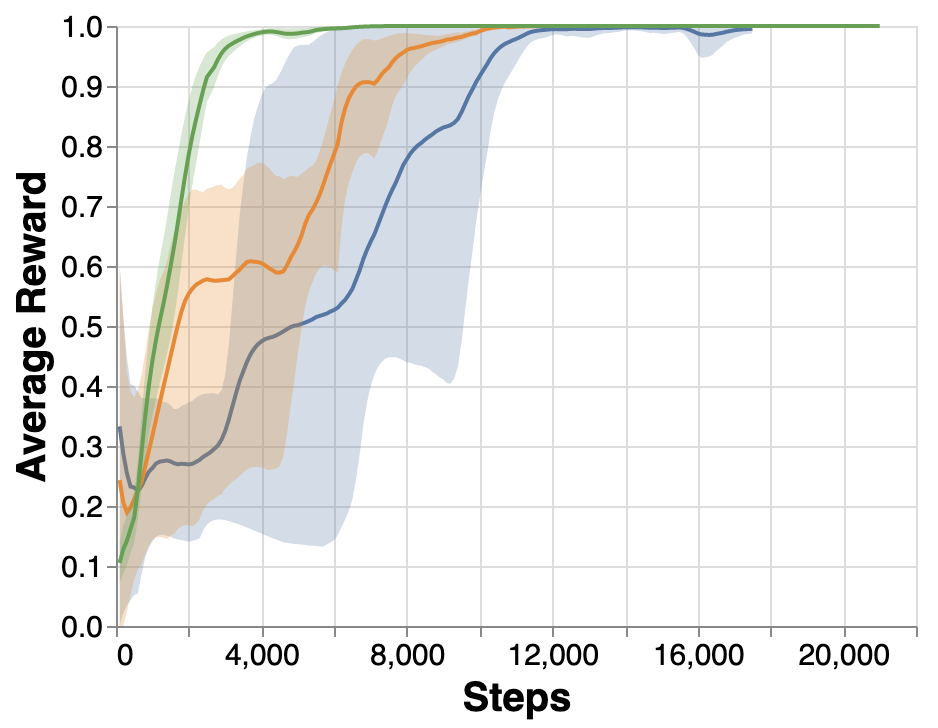}
    \caption{X-Island (1): Average reward per episode}
    \label{fig:xisland_reward_ppo_classifier}
\end{subfigure}
\begin{subfigure}[t]{0.3\textwidth}\centering
    \includegraphics[width=\textwidth]{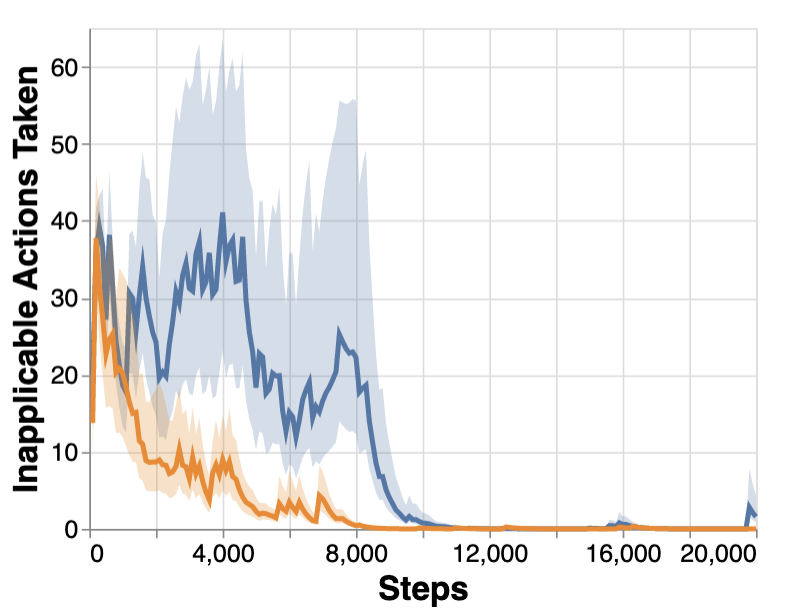}
    \caption{X-Island (1): Average number of inapplicable actions taken per episode}
    \label{fig:xisland_inapplicable_actions_ppo_classifier}
\end{subfigure}
\begin{subfigure}[t]{0.3\textwidth}\centering
    \includegraphics[width=\textwidth]{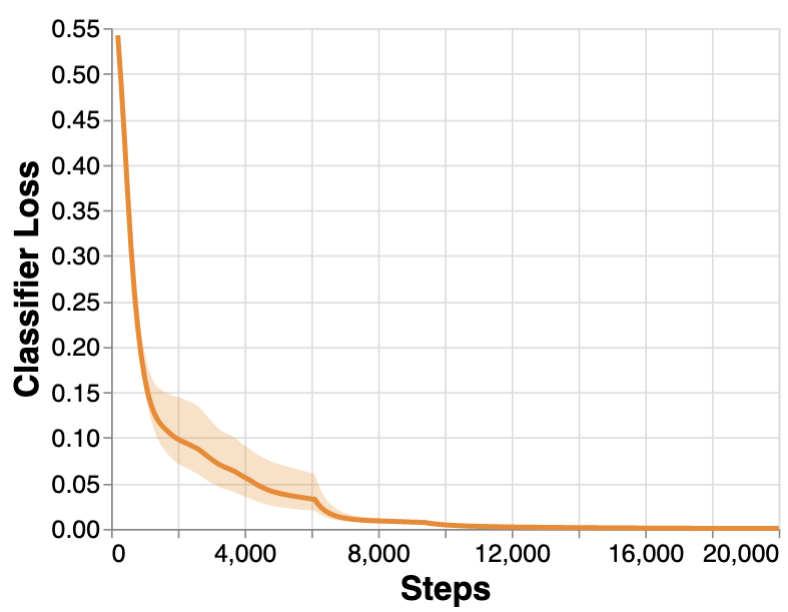}
    \caption{X-Island (1): Classifier loss}
    \label{fig:xisland_classifier_loss_ppo_classifier}
\end{subfigure}

\begin{subfigure}[t]{0.3\textwidth}\centering
    \includegraphics[width=\textwidth]{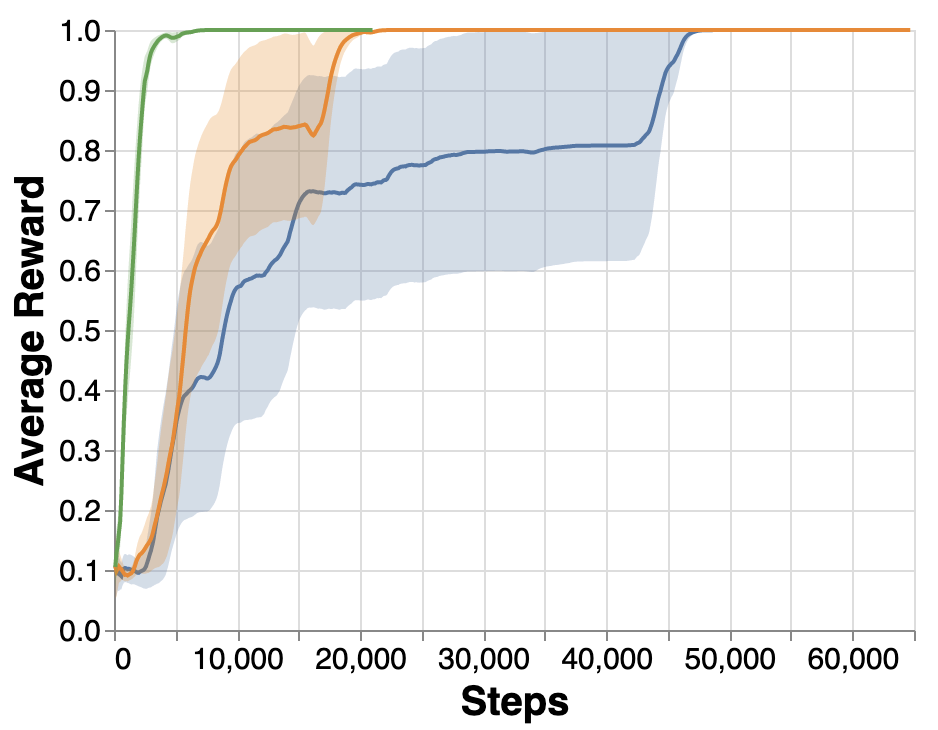}
    \caption{X-Island (2): Average reward per episode}
    \label{fig:xislandbis_reward_ppo_classifier}
\end{subfigure}
\begin{subfigure}[t]{0.3\textwidth}\centering
    \includegraphics[width=\textwidth]{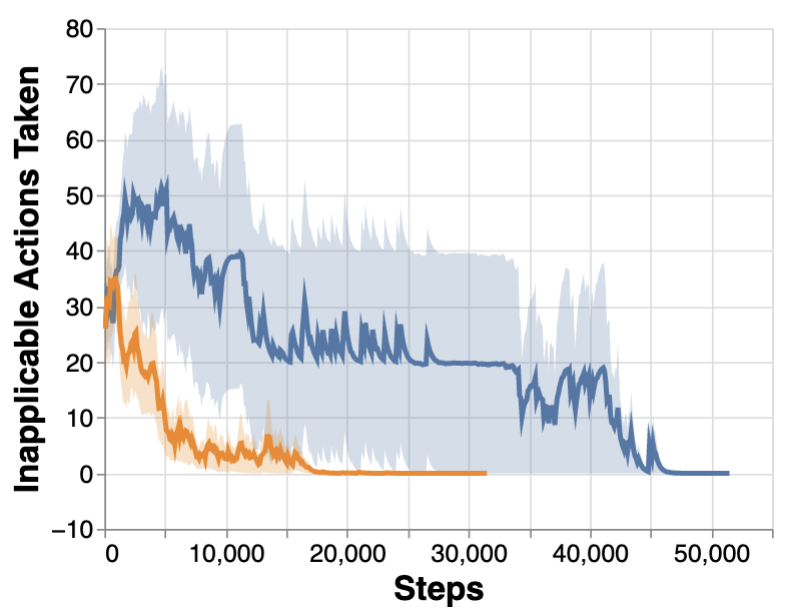}
    \caption{X-Island (2): Average number of inapplicable actions taken per episode}
    \label{fig:xislandbis_inapplicable_actions_ppo_classifier}
\end{subfigure}
\begin{subfigure}[t]{0.3\textwidth}\centering
    \includegraphics[width=\textwidth]{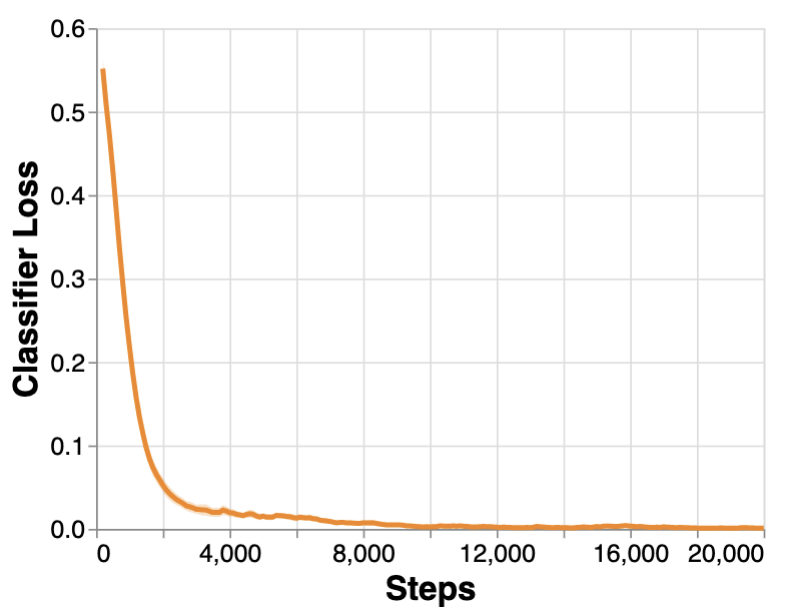}
    \caption{X-Island (2): Classifier loss}
    \label{fig:xislandbis_classifier_loss_ppo_classifier}
\end{subfigure}

\begin{subfigure}[t]{0.3\textwidth}\centering
    \includegraphics[width=\textwidth]{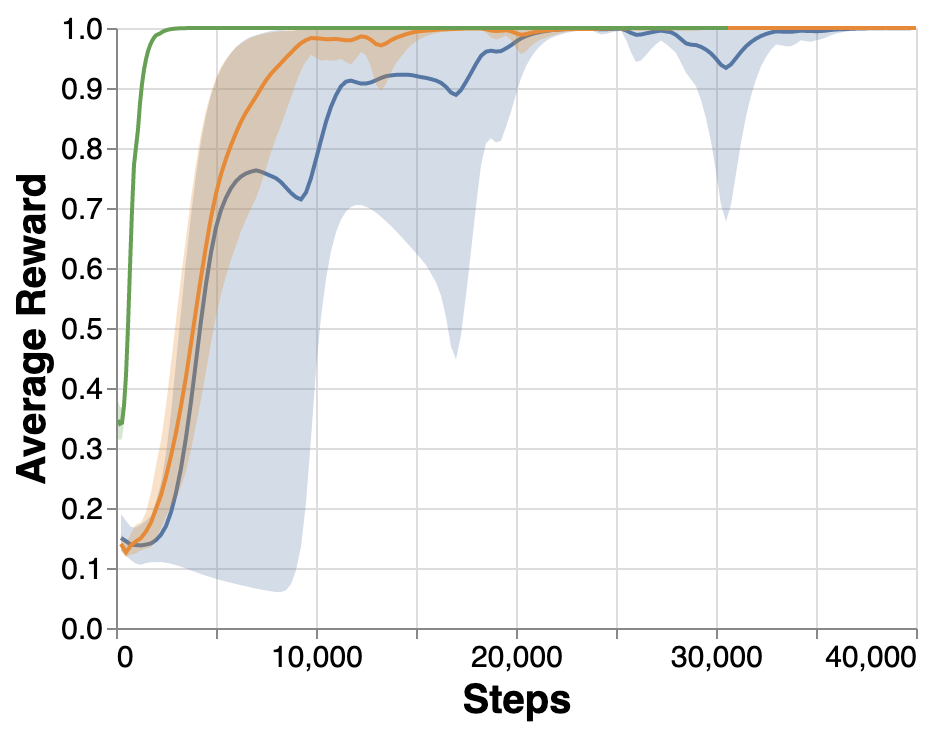}
    \caption{Key \& Door (1): Average reward per episode}
    \label{fig:doorkey_reward_ppo_classifier}
\end{subfigure}
\begin{subfigure}[t]{0.3\textwidth}\centering
    \includegraphics[width=\textwidth]{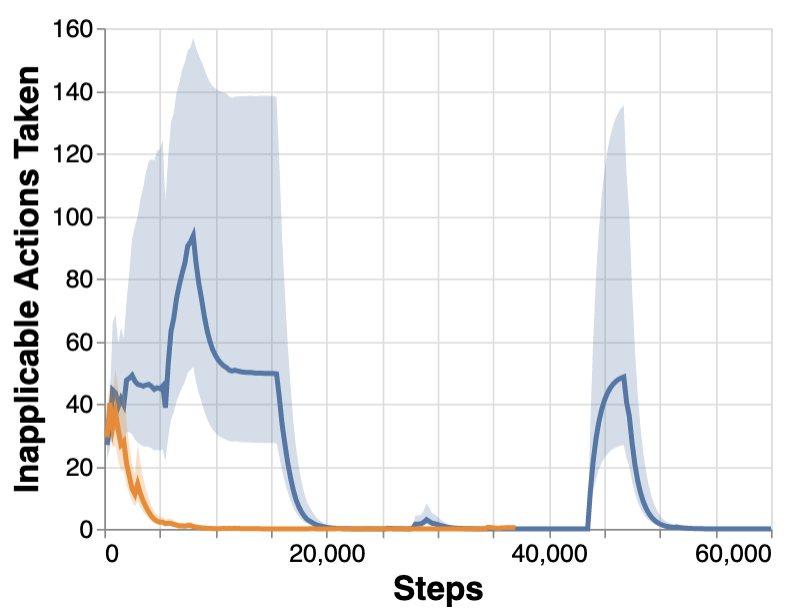}
    \caption{Key \& Door (1): Average number of inapplicable actions taken per episode}
    \label{fig:doorkey_inapplicable_actions_ppo_classifier}
\end{subfigure}
\begin{subfigure}[t]{0.3\textwidth}\centering
    \includegraphics[width=\textwidth]{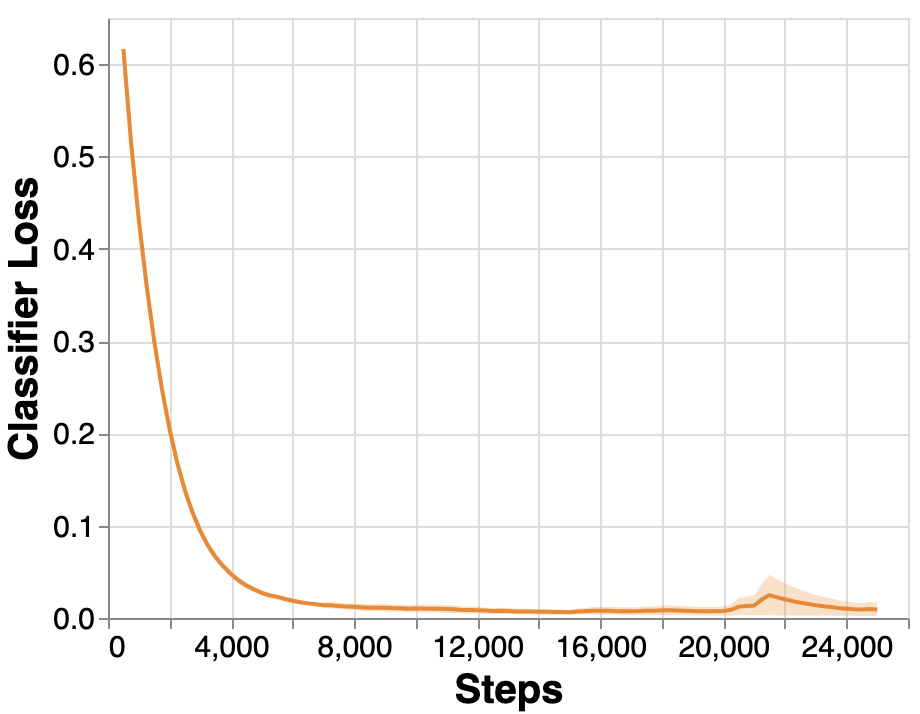}
    \caption{Key \& Door (1): Classifier loss}
    \label{fig:doorkey_classifier_loss_ppo_classifier}
\end{subfigure}

\begin{subfigure}[t]{0.3\textwidth}\centering
    \includegraphics[width=\textwidth]{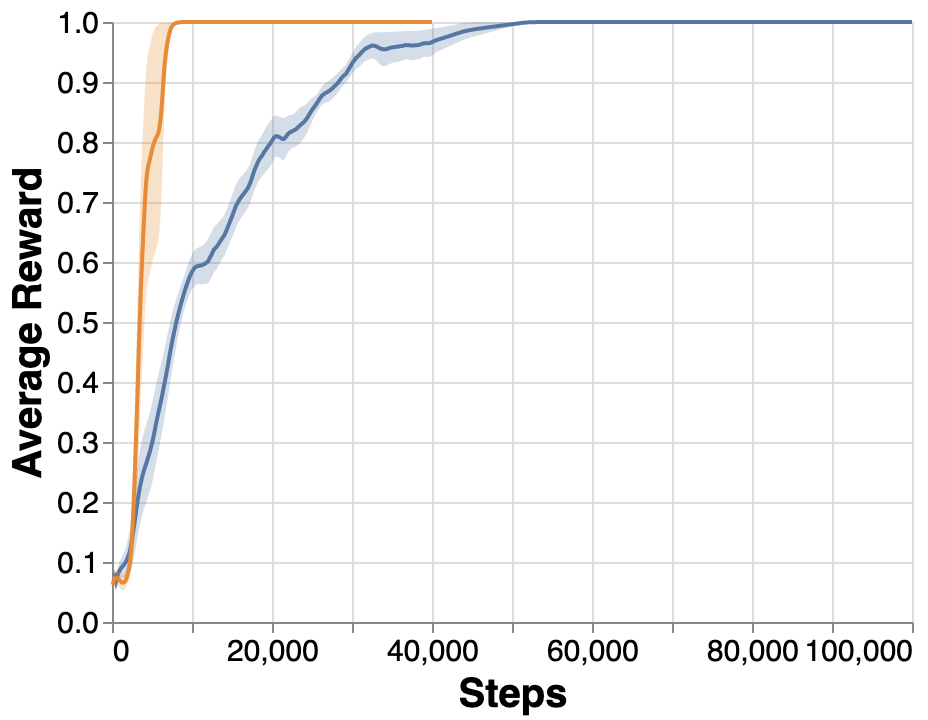}
    \caption{Key \& Door (2): Average reward per episode}
    \label{fig:doorkey2_reward_ppo_classifier}
\end{subfigure}
\begin{subfigure}[t]{0.3\textwidth}\centering
    \includegraphics[width=\textwidth]{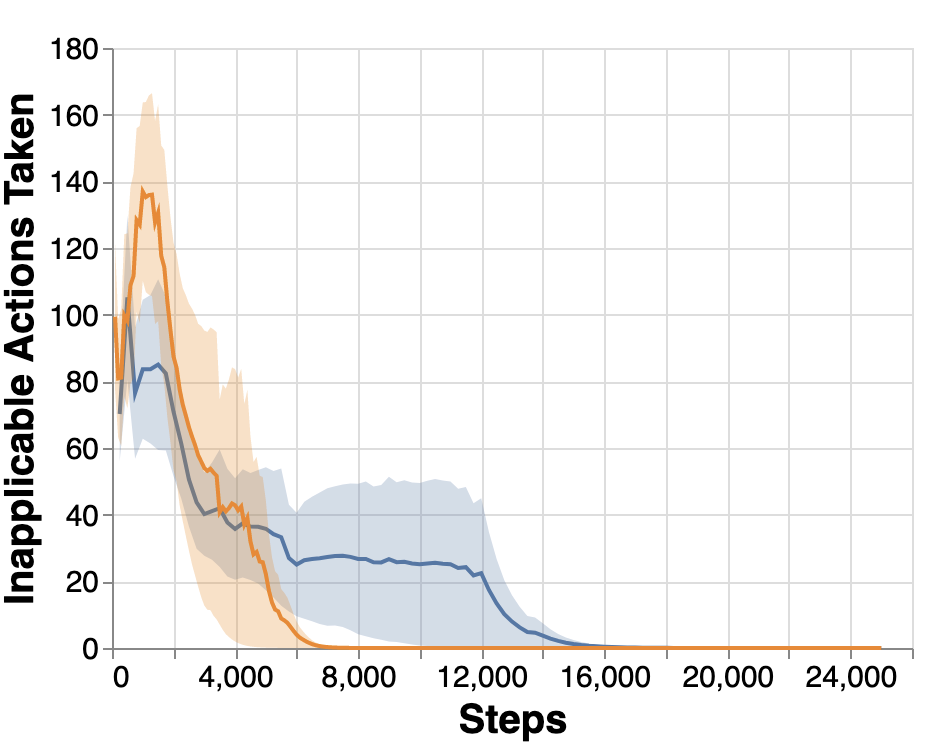}
    \caption{Key \& Door (2): Average number of inapplicable actions taken per episode}
    \label{fig:doorkey2_inapplicable_actions_ppo_classifier}
\end{subfigure}
\begin{subfigure}[t]{0.3\textwidth}\centering
    \includegraphics[width=\textwidth]{figures/doorkey_classifier_loss_ppo_vs_classifier.png}
    \caption{Key \& Door (2): Classifier loss}
    \label{fig:doorkey2_classifier_loss_ppo_classifier}
\end{subfigure}

\caption{Inapplicable Actions Learning}
\label{fig:xisland_ppo_classifier}
\end{figure*}

Introducing the inapplicable actions learning classifier helps reduce the sample complexity of the PPO algorithm in all the tasks \figref{fig:xisland_ppo_classifier}. The classifier is able to quickly learn to identify the inapplicable actions, reducing the number of inapplicable actions explored.

\subsection{Inapplicable actions learning with partial domain knowledge}
\label{sec:partial_plus_classifier}

\begin{figure}[H]
    \centering
    \includegraphics[width=0.3\textwidth]{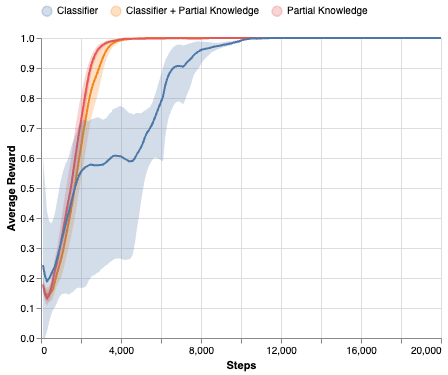}
    \caption{Average reward for different combinations of Classifier and Partial Knowledge.}
    \label{fig:x_island_partial_vs_scratch}
\end{figure}

As we can see in Figure~\ref{fig:x_island_partial_vs_scratch}, in this case jointly using partial knowledge and the classifier to learn the rest of the applicability function $C$ perform similarly, both outperforming learning the classifier from scratch.
This is because learning the classifier takes a similar time than learning the optimal policy in this particular task.

\section{Classifier analysis} \label{sec:classifier_analysis_additional}

\begin{figure}[H]
\centering
\captionsetup{justification=centering}
\captionsetup[sub]{font=scriptsize}
\captionsetup[sub]{justification=centering}
\hspace{\fill}
\begin{subfigure}[t]{0.1\textwidth}\centering
    \includegraphics[width=\textwidth]{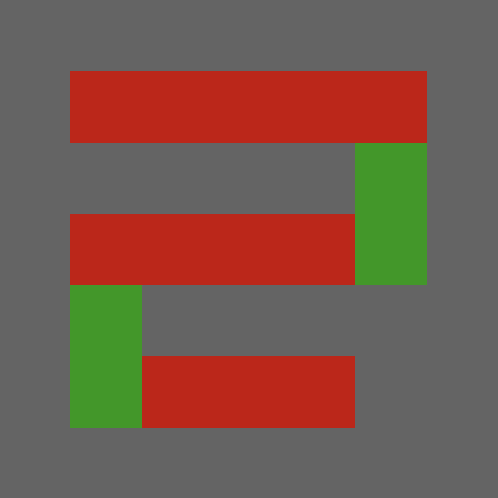}
    \caption{Action \texttt{up}}
    \label{fig:maze_classifier_up}
\end{subfigure}
\hfill
\begin{subfigure}[t]{0.1\textwidth}\centering
    \includegraphics[width=\textwidth]{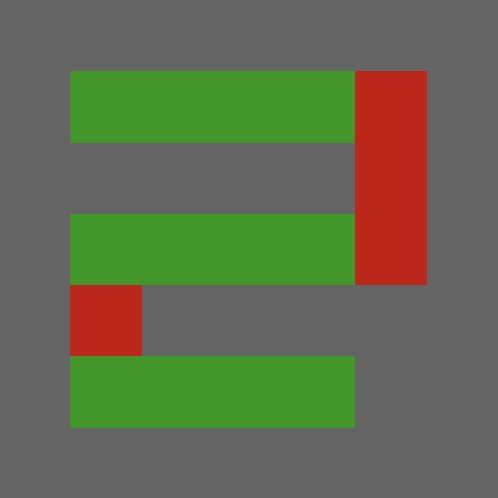}
    \caption{Action \texttt{right}}
    \label{fig:maze_classifier_right}
\end{subfigure}
\hfill
\begin{subfigure}[t]{0.1\textwidth}\centering
    \includegraphics[width=\textwidth]{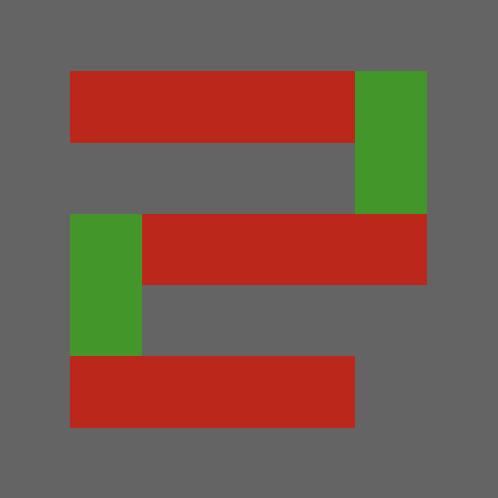}
    \caption{Action \texttt{down}}
    \label{fig:maze_classifier_down}
\end{subfigure}
\hfill
\begin{subfigure}[t]{0.1\textwidth}\centering
    \includegraphics[width=\textwidth]{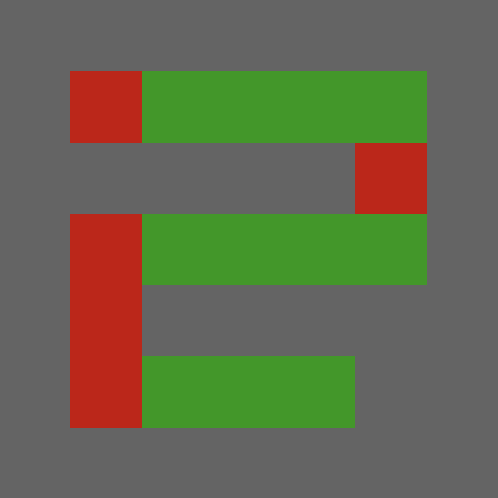}
    \caption{Action \texttt{left}}
    \label{fig:maze_classifier_left}
\end{subfigure}
\hspace{\fill}

\hspace{\fill}
\begin{subfigure}[t]{0.1\textwidth}\centering
    \includegraphics[width=\textwidth]{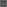}
    \caption{Action \texttt{up}}
    \label{fig:xislandbis_classifier_up}
\end{subfigure}
\hfill
\begin{subfigure}[t]{0.1\textwidth}\centering
    \includegraphics[width=\textwidth]{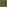}
    \caption{Action \texttt{right}}
    \label{fig:xislandbis_classifier_right}
\end{subfigure}
\hfill
\begin{subfigure}[t]{0.1\textwidth}\centering
    \includegraphics[width=\textwidth]{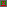}
    \caption{Action \texttt{down}}
    \label{fig:xislandbis_classifier_down}
\end{subfigure}
\hfill
\begin{subfigure}[t]{0.1\textwidth}\centering
    \includegraphics[width=\textwidth]{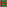}
    \caption{Action \texttt{left}}
    \label{fig:xislandbis_classifier_left}
\end{subfigure}
\hspace{\fill}

\hspace{\fill}
\begin{subfigure}[t]{0.1\textwidth}\centering
    \includegraphics[width=\textwidth]{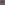}
    \caption{Action \texttt{up}}
    \label{fig:doorkey_classifier_up}
\end{subfigure}
\hfill
\begin{subfigure}[t]{0.1\textwidth}\centering
    \includegraphics[width=\textwidth]{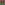}
    \caption{Action \texttt{right}}
    \label{fig:doorkey_classifier_right}
\end{subfigure}
\hfill
\begin{subfigure}[t]{0.1\textwidth}\centering
    \includegraphics[width=\textwidth]{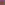}
    \caption{Action \texttt{down}}
    \label{fig:doorkey_classifier_down}
\end{subfigure}
\hfill
\begin{subfigure}[t]{0.1\textwidth}\centering
    \includegraphics[width=\textwidth]{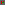}
    \caption{Action \texttt{left}}
    \label{fig:doorkey_classifier_left}
\end{subfigure}
\hspace{\fill}

\hspace{\fill}
\begin{subfigure}[t]{0.1\textwidth}\centering
    \includegraphics[width=\textwidth]{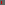}
    \caption{Action \texttt{right} when door is closed}
    \label{fig:doorkey_classifier_right_no_door}
\end{subfigure}
\hfill
\begin{subfigure}[t]{0.1\textwidth}\centering
    \includegraphics[width=\textwidth]{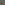}
    \caption{Action \texttt{pickup}}
    \label{fig:doorkey_classifier_pickup}
\end{subfigure}
\hfill
\begin{subfigure}[t]{0.1\textwidth}\centering
    \includegraphics[width=\textwidth]{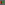}
    \caption{Action \texttt{open} when door is open}
    \label{fig:doorkey_classifier_open}
\end{subfigure}
\hspace{\fill}

\caption{Classifier output learnt in the Maze, X-Island (2) and Key \& Door environments}
\label{fig:classifier_additional}
\end{figure}

\figref{fig:classifier_additional} presents the output of the classifier trained in the different environments. The classifier is able to predict with high accuracy that an action is inapplicable in a given state in all the environments. In the Key \& Door environment, the applicability of the action \texttt{right} changes when the door is open. We see in Figures~\ref{fig:doorkey_classifier_right_no_door} and~\ref{fig:doorkey_classifier_right} that the classifier is able to capture the change in the state that makes an action applicable. The classifier struggles however with the \texttt{open} action for states that were not visited often \figref{fig:doorkey_classifier_open}.


\section{Exploration Parameter Analysis} 

\begin{figure}[H]
    \centering
    \captionsetup{justification=centering}
    \includegraphics[width=0.3\textwidth]{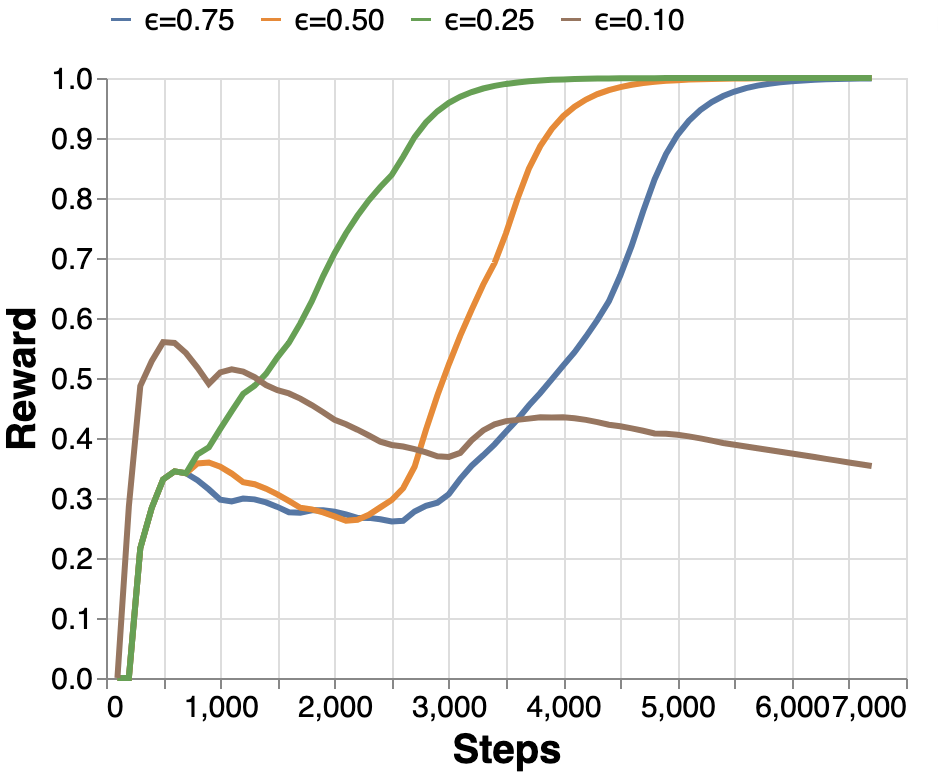}
    \caption{Reward for different exploration parameter values ($\epsilon$) in the Maze Environment.}
    \label{fig:epsilon_parameter}
\end{figure}

To understand its impact on the sample efficiency, we vary the value of the exploration parameter $\epsilon$ of \algoref{alg:algorithm} while learning an optimal policy in the Maze domain. The exploration parameter represents the probability at which the mask generated by the inapplicable action classifier will be ignored (\algolineref{alg_line:random_use_classifier} of \algoref{alg:algorithm}).

We observe in \figref{fig:epsilon_parameter} that a trade-off exists between exploration and exploitation of the knowledge acquired about inapplicable actions. When the agent explores too much ($\epsilon=0.75$) the agent needs more time to converge to the optimal policy. With a lower $\epsilon$, the agent is able to reduce the time to converge by leveraging the accuracy of the classifier to mask inapplicable actions. However, when the agent relies too much on the classifier ($\epsilon=0.1$) and therefore does not explore enough the valid actions as the classifier incorrectly masks them, making the reward diverge.

\section{Knowledge transfer} \label{sec:knowledge_transfer_experiments_additional}

\textbf{PPO Warm Start}: PPO Warm Start reuses a pre-trained policy as a starting point to learn a new task. The weights of the original policy are used to instantiate the current policy before the training starts.

\textbf{PPO Reuse}: We implement the logic of Policy Reuse ~\citep{Fernandez_Garcia_Veloso_2010} in the PPO algorithm, where during the rollouts, a pre-trained expert policy is sampled with a probability $\epsilon$ and the current policy is used the rest of the time.

\begin{figure*}[h]
\begin{minipage}{\textwidth}
\centering
\captionsetup{justification=centering}

\hspace{\fill}
\begin{subfigure}[t]{0.45\textwidth}\centering
    \includegraphics[width=\textwidth]{figures/transfer/legend.png}
\end{subfigure}
\hspace{\fill}


\begin{subfigure}[t]{0.24\textwidth}\centering
    \includegraphics[width=\textwidth]{figures/transfer/maze_reuse_maze.png}
\end{subfigure}
\begin{subfigure}[t]{0.23\textwidth}\centering
    \includegraphics[width=\textwidth]{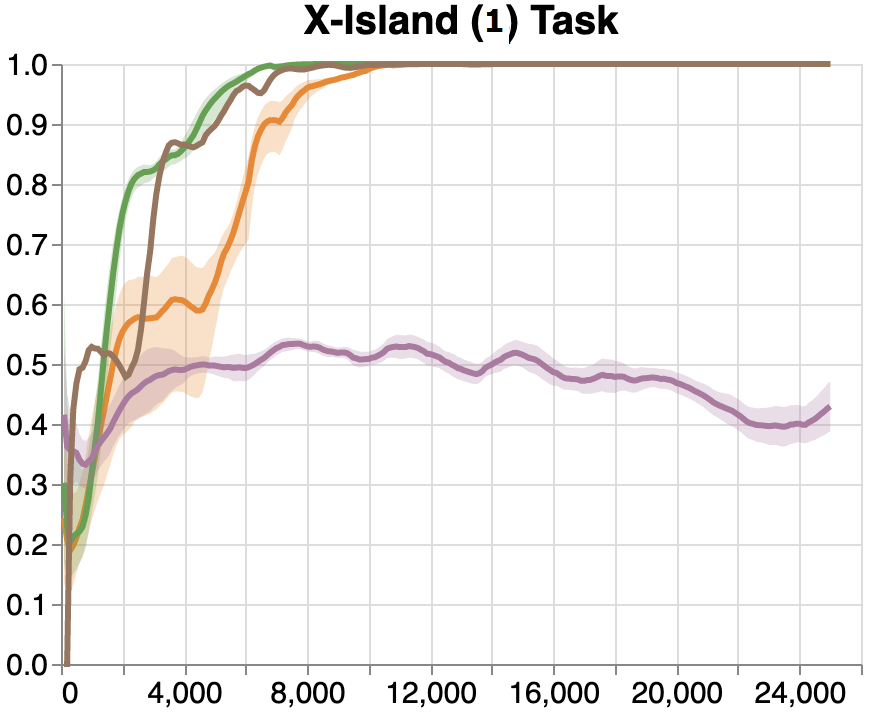}
\end{subfigure}
\begin{subfigure}[t]{0.23\textwidth}\centering
    \includegraphics[width=\textwidth]{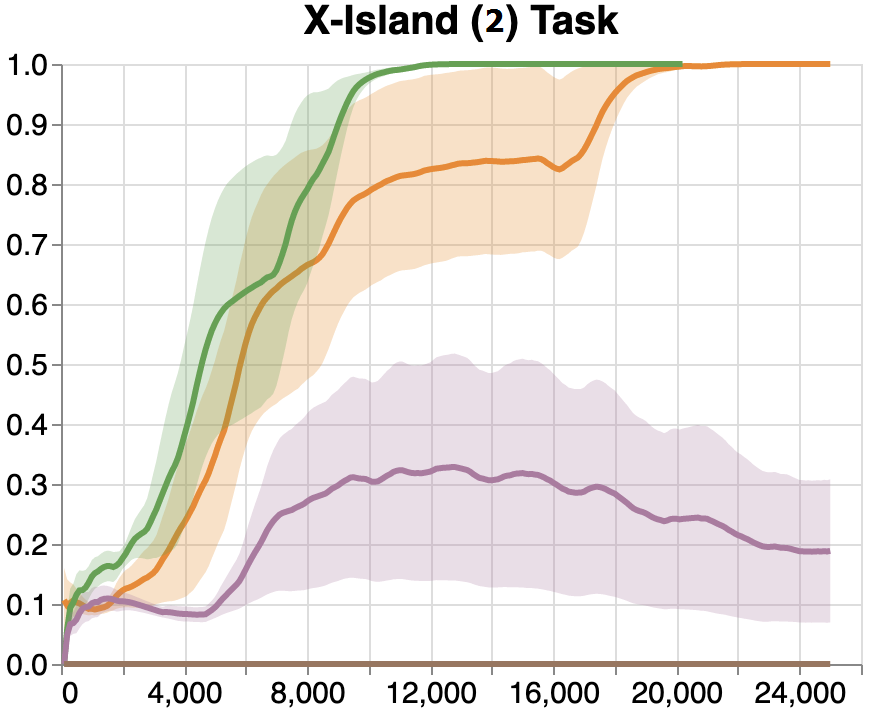}
\end{subfigure}
\begin{subfigure}[t]{0.23\textwidth}\centering
    \includegraphics[width=\textwidth]{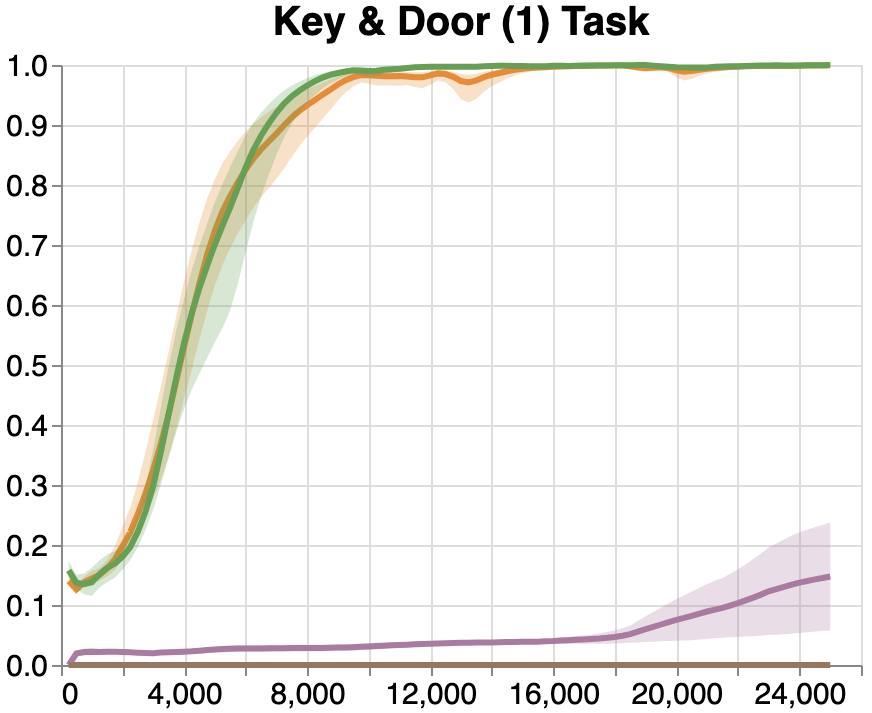}
\end{subfigure}

\begin{subfigure}[t]{0.24\textwidth}\centering
    \includegraphics[width=\textwidth]{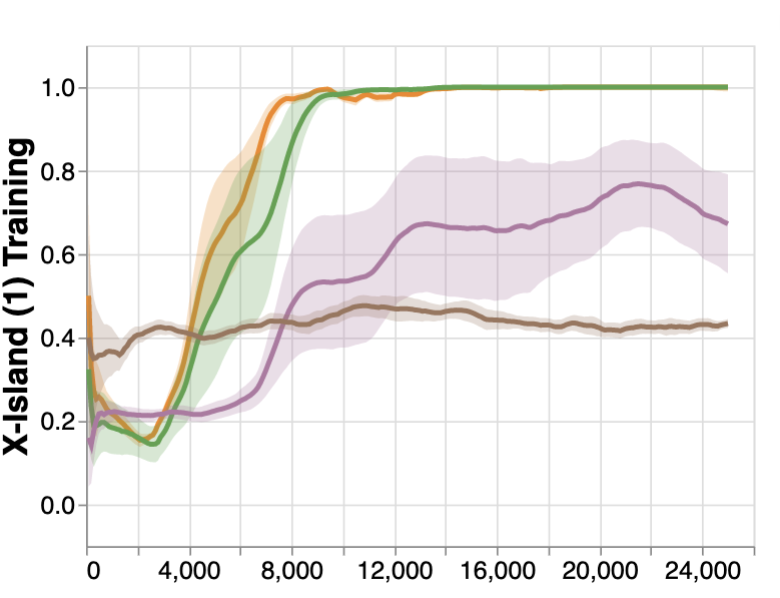}
\end{subfigure}
\begin{subfigure}[t]{0.24\textwidth}\centering
    \includegraphics[width=\textwidth]{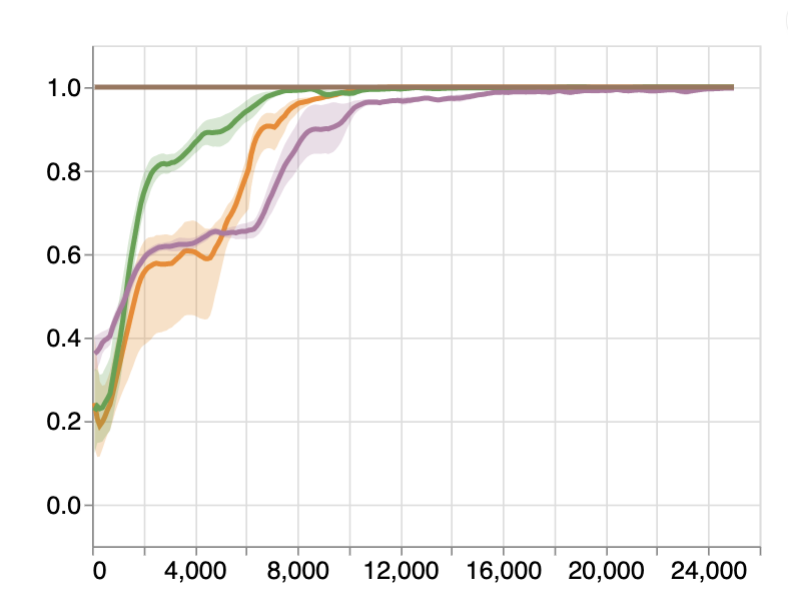}
\end{subfigure}
\begin{subfigure}[t]{0.24\textwidth}\centering
    \includegraphics[width=\textwidth]{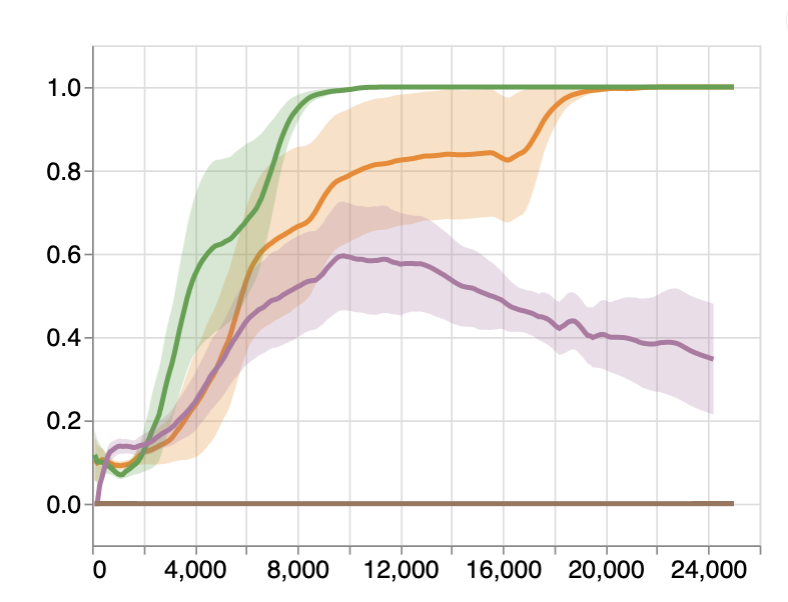}
\end{subfigure}
\begin{subfigure}[t]{0.24\textwidth}\centering
    \includegraphics[width=\textwidth]{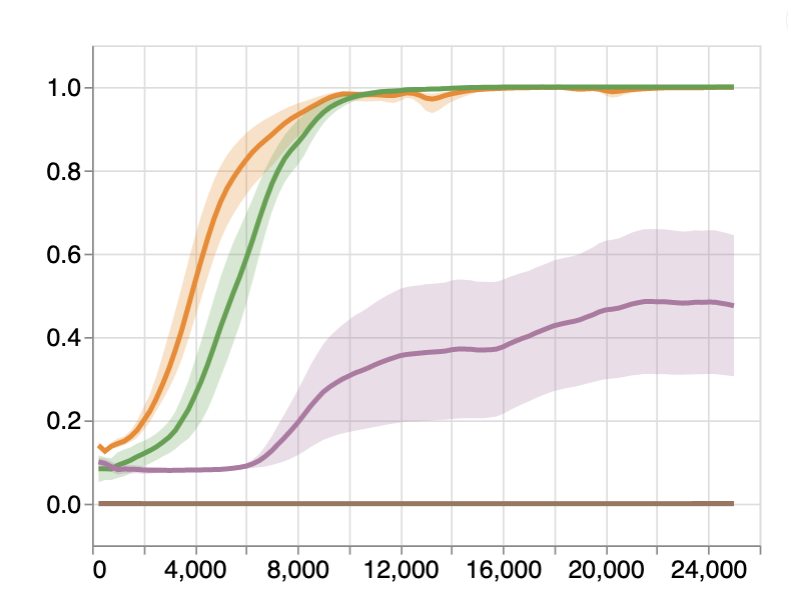}
\end{subfigure}

\begin{subfigure}[t]{0.24\textwidth}\centering
    \includegraphics[width=\textwidth]{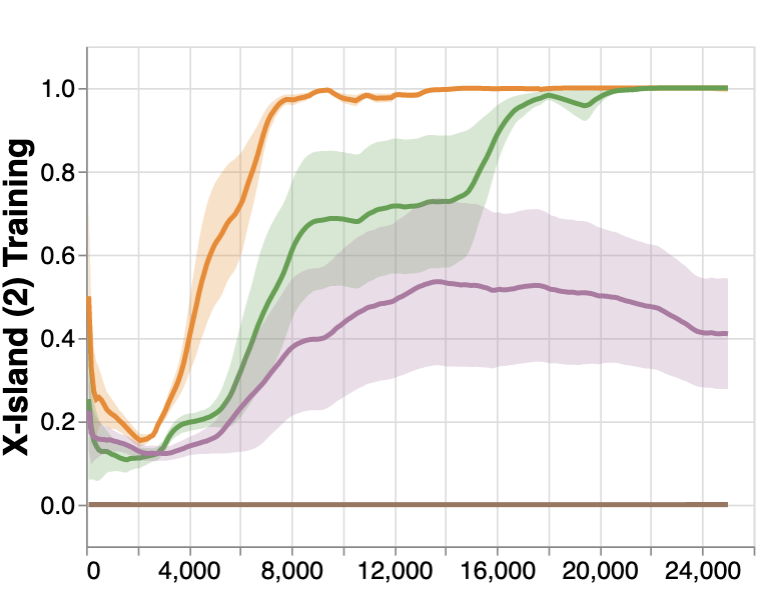}
\end{subfigure}
\begin{subfigure}[t]{0.24\textwidth}\centering
    \includegraphics[width=\textwidth]{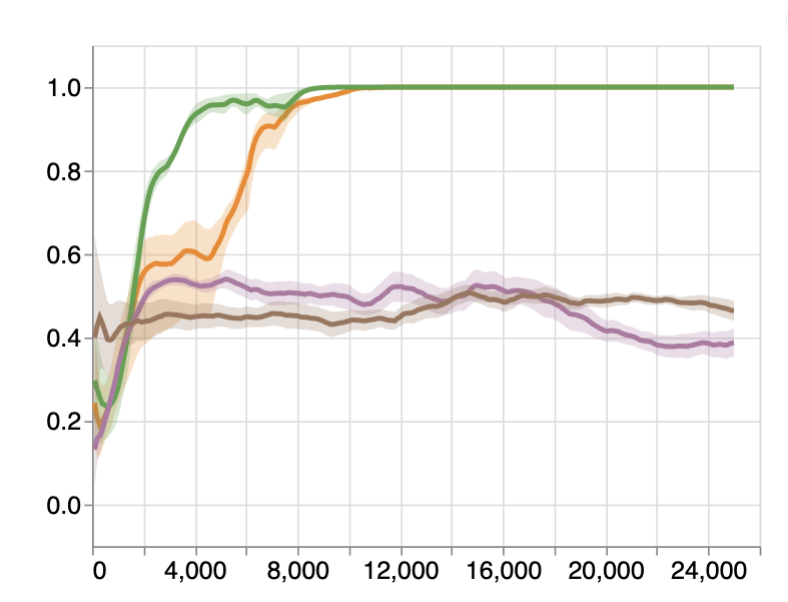}
\end{subfigure}
\begin{subfigure}[t]{0.24\textwidth}\centering
    \includegraphics[width=\textwidth]{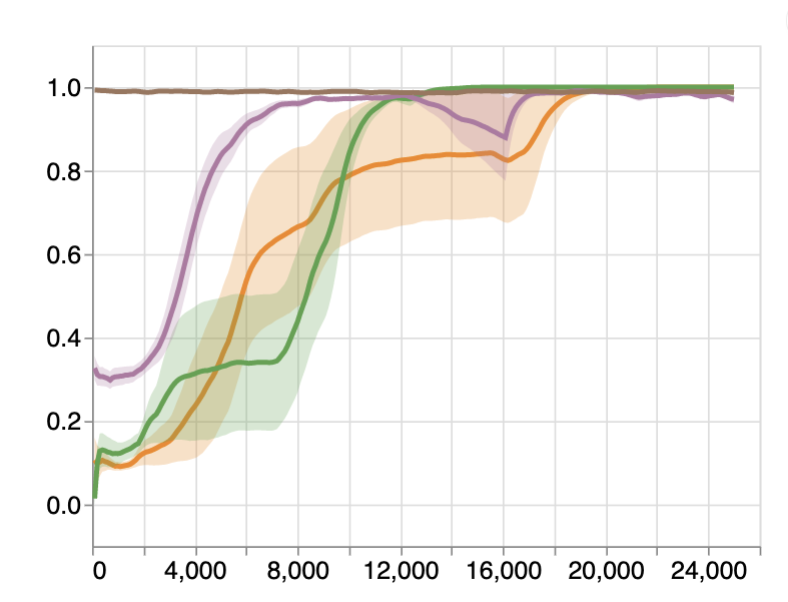}
\end{subfigure}
\begin{subfigure}[t]{0.24\textwidth}\centering
    \includegraphics[width=\textwidth]{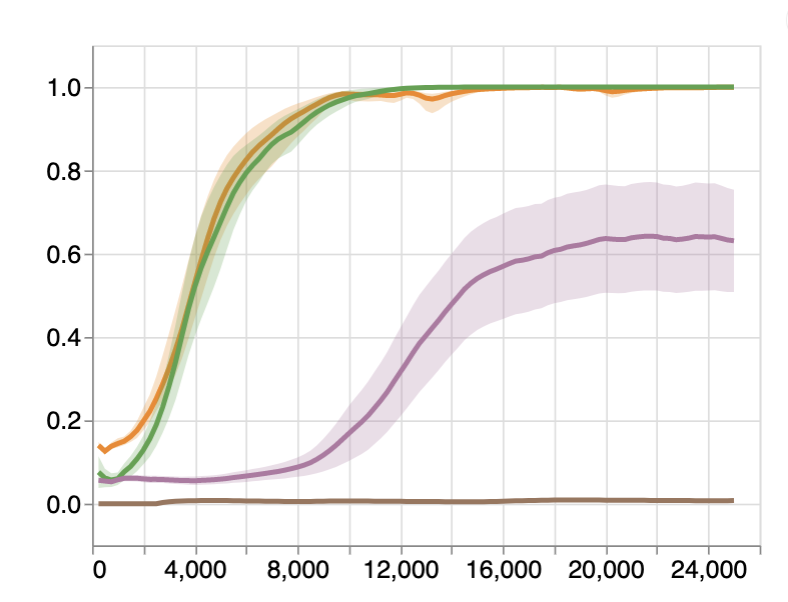}
\end{subfigure}

\begin{subfigure}[t]{0.24\textwidth}\centering
    \includegraphics[width=\textwidth]{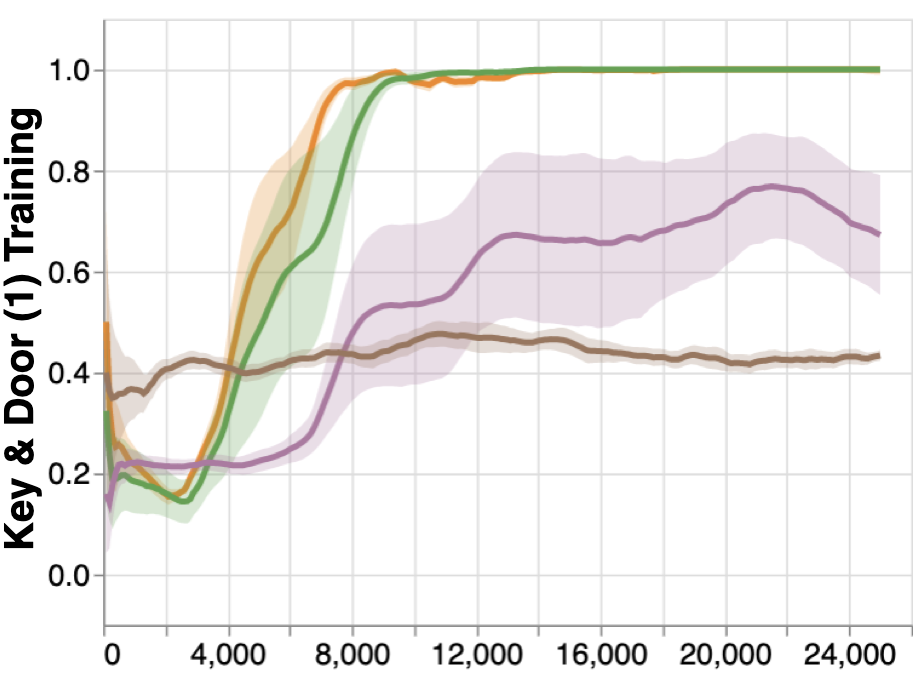}
\end{subfigure}
\begin{subfigure}[t]{0.24\textwidth}\centering
    \includegraphics[width=\textwidth]{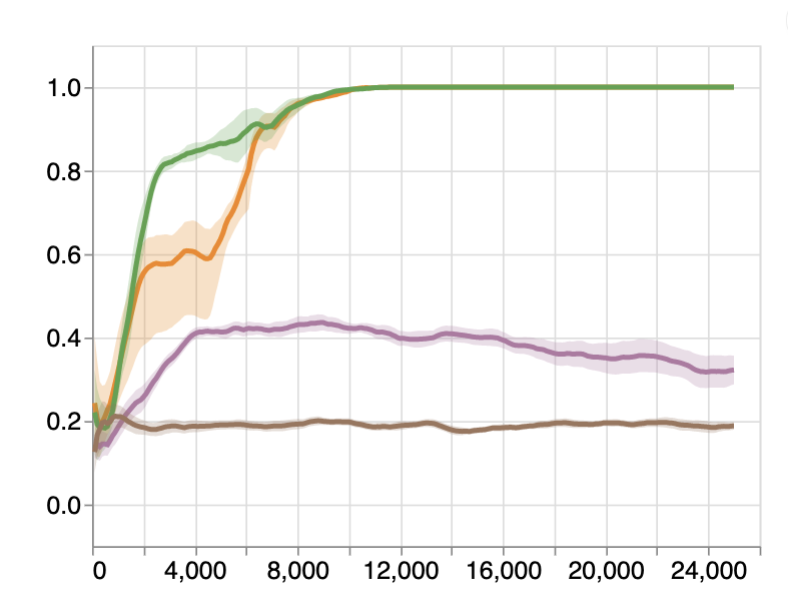}
\end{subfigure}
\begin{subfigure}[t]{0.24\textwidth}\centering
    \includegraphics[width=\textwidth]{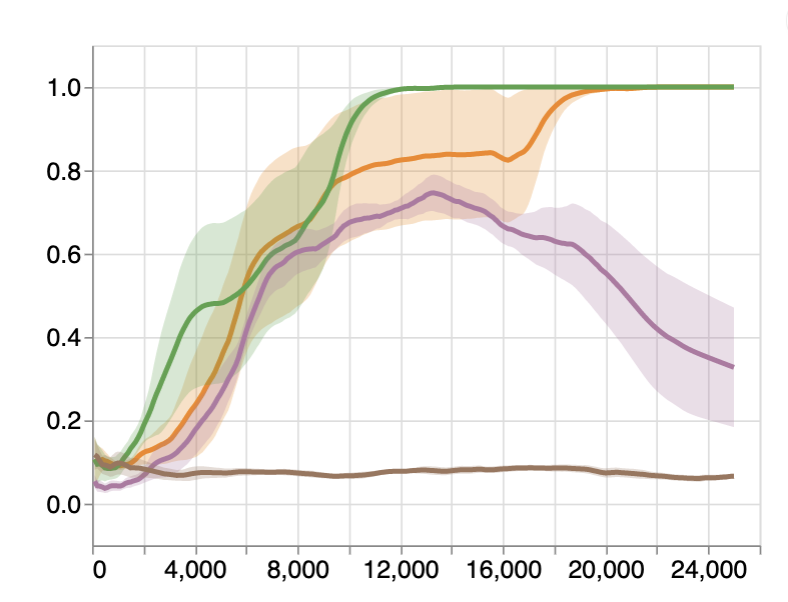}
\end{subfigure}
\begin{subfigure}[t]{0.24\textwidth}\centering
    \includegraphics[width=\textwidth]{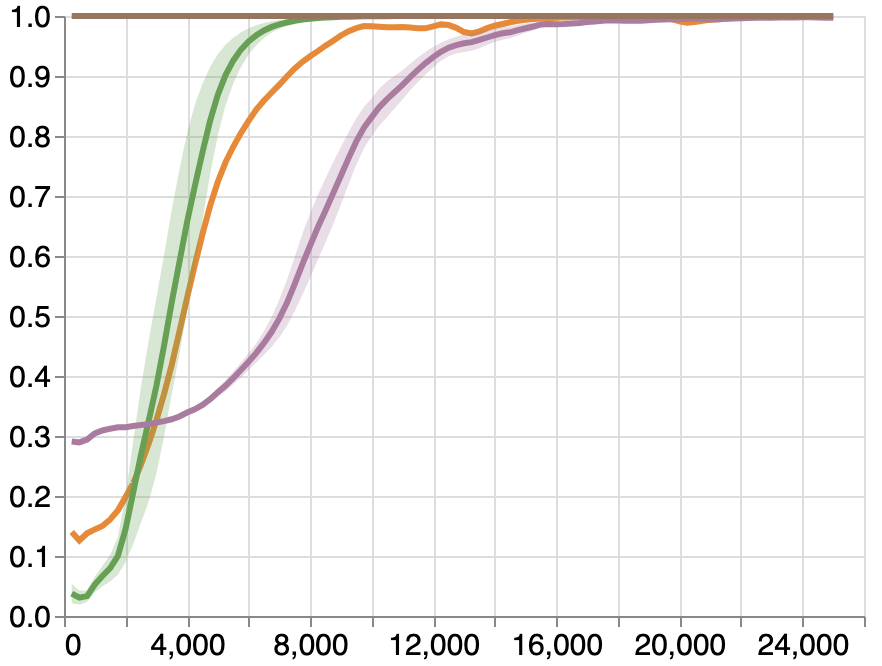}
\end{subfigure}

\caption{Average reward for different techniques of knowledge transfer reusing the knowledge acquired in the environment specified as the row title and applied to the environment specified as the column title.}
\label{fig:classifier_reuse_additional}
\end{minipage}
\end{figure*}

As observed in \figref{fig:classifier_reuse_additional}, the transfer of the trained classifier from one environment to another is beneficial. The training efficiency is improved in almost all the configurations. The performance improvement is more important when the training environment and the destination environment are similar. A more constrained environment, such as the Maze environment with only 50\% of the action space being applicable in most of the states, seems to provide better knowledge transfer results. Comparing the transfer of the classifier with other knowledge transfer techniques such as Policy Reuse or a naive warm start approach, our approach is more versatile. It can be transferred both across tasks and across environments. The warm start approach, clearly overfits while the Policy Reuse learns something but struggles to find an optimal policy.

\end{document}